\documentclass{article}

\usepackage[preprint]{neurips_2026}

\usepackage[utf8]{inputenc} 
\usepackage[T1]{fontenc}    
\usepackage{hyperref}       
\usepackage{url}            
\usepackage{booktabs}       
\usepackage{amsfonts}       
\usepackage{nicefrac}       
\usepackage{microtype}      
\usepackage{xcolor}         
\usepackage{adjustbox}
\usepackage{algorithm}
\usepackage{algpseudocode}
\usepackage{booktabs}
\usepackage{graphicx}
\usepackage{longtable}
\usepackage{lineno}
\usepackage{CJKutf8}
\usepackage{colortbl}
\usepackage{capt-of}
\usepackage{enumitem}
\usepackage{caption}
\usepackage{comment}
\usepackage{siunitx}
\usepackage{amsmath}

\usepackage{textcomp}
\usepackage[most]{tcolorbox}
\tcbuselibrary{breakable,skins,listings}

\lstdefinestyle{promptstyle}{
  basicstyle=\ttfamily\small,
  breaklines=true,
  breakatwhitespace=false,
  breakindent=0pt,
  breakautoindent=false,
  columns=fullflexible,
  keepspaces=true,
  showstringspaces=false,
  upquote=true,
  literate=
    {–}{{\textendash}}1
    {—}{{\textemdash}}1
    {“}{{``}}1
    {”}{{''}}1
    {‘}{{`}}1
    {’}{{'}}1
    {…}{{\ldots}}1
    {→}{{$\to$}}1
    {←}{{$\leftarrow$}}1
    {≥}{{$\ge$}}1
    {≤}{{$\le$}}1
    {≠}{{$\ne$}}1
    {÷}{{$\div$}}1
}

\newtcblisting{promptbox}[2][]{%
  enhanced,
  breakable,
  listing only,
  listing style=promptstyle,
  colback=gray!4,
  colframe=black!55,
  coltitle=black,
  colbacktitle=gray!10,
  fonttitle=\bfseries,
  title={#2},
  boxrule=0.6pt,
  arc=2pt,
  left=6pt,
  right=6pt,
  top=6pt,
  bottom=6pt,
  #1
}

\title{The Illusion of Robustness: Aggregate Accuracy Hides Prediction Flips under Task-Irrelevant Context}

\author{%
  Yanzhe Zhang \\
  Georgia Tech \\
  \texttt{z\_yanzhe@gatech.edu} \\
  \And
  Sanmi Koyejo \\
  Stanford University \\
  \texttt{sanmi@stanford.edu} \\
  \And
  Diyi Yang\\
  Stanford University \\
  \texttt{diyiy@stanford.edu} \\
}

\begin{document}

\maketitle

\begin{abstract}
As large language models (LLMs) grow more capable, they are increasingly deployed in context-rich settings where task inputs are often accompanied by long, partially irrelevant context.
In a controlled setting, we find that state-of-the-art models often appear robust to task-irrelevant context at the aggregate level: prepending it to benchmark questions causes little change in overall accuracy. This aggregate stability, however, masks significant per-example instability. Even semantically meaningless pseudo-words, formed by randomly combining characters, can markedly shift model predictions on a small fraction of examples, degrading performance on some while improving it on others.
This two-sided effect holds consistently across a wide range of models and datasets, yet the affected examples are largely model-specific. We further show that this instability is modulated by context type, context length, test-time compute, and model development stage. Together, our findings reveal context-induced tail risks concealed by aggregate accuracy, motivating per-example reliability evaluation of language models.\footnote{Code and data are available at \url{https://github.com/SALT-NLP/illusion-of-robustness}.}
\end{abstract}

\section{Introduction}

State-of-the-art large language models (LLMs) are increasingly deployed as agents operating in context-rich environments, where they rarely receive a clean task input in isolation.
A coding agent \citep{jimenez2024swebenchlanguagemodelsresolve, yang2024sweagentagentcomputerinterfacesenable} may need to inspect long source files, execution logs, and intermediate tool outputs before making a small code edit.
A deep research agent \citep{OpenAIDeepResearch, zheng-etal-2025-deepresearcher} may condition on noisy web retrieval results before answering a factual question.
A computer-use agent \citep{xie2024osworld, wang2025opencuaopenfoundationscomputeruse} may process a long sequence of screenshots before deciding its next step.
In all of these scenarios, the task-relevant signal is embedded within a much larger amount of task-irrelevant context.

In this work, we revisit the common assumption that irrelevant context should not substantially alter a model's answer. We use a simplified setting to isolate the effect of task-irrelevant context: we prepend it to questions from QA benchmarks and measure the change in performance.

At the aggregate level, state-of-the-art models appear robust: benchmark-level accuracy changes by no more than \(\pm0.9\%\) for \texttt{gpt-5.5} and \(\pm2.1\%\) for \texttt{gpt-5.4} after adding task-irrelevant context.
However, this apparent robustness hides substantial per-example instability, as the underlying phenomenon is two-sided: performance drops sharply on some questions, while others become ``easier'' and show large gains, even though no additional task-relevant information is provided. We refer to such prediction flips as \textbf{context-induced instability}.

This instability matters in real-world deployments because a shift at a single step may trigger an irreversible action or propagate through a sequence of downstream decisions.
To characterize the phenomenon, we introduce two complementary metrics.
\emph{Instability (INS)} measures the average absolute change in performance across all examples, capturing how sensitive a model is to irrelevant context regardless of whether the change helps or hurts.
\emph{Worst-tail Degradation (WTD)}, in contrast, measures the average performance drop within the worst-affected tail, capturing the risk of severe context-induced failures.

We observe this phenomenon across different models on multiple datasets (MMLU-Pro \citep{wang2024mmluprorobustchallengingmultitask}, GPQA \citep{rein2023gpqagraduatelevelgoogleproofqa}, Humanity's Last Exam (HLE)  \citep{phan2025humanity}, and SimpleQA \citep{wei2024measuringshortformfactualitylarge}), where INS reaches $13.6\%$ and WTD reaches $53.2\%$.
Although the phenomenon itself is consistent, the specific examples it affects are model-specific rather than simply data-dependent: per-example performance shifts are poorly correlated across models (e.g., on MMLU-Pro, the mean Pearson $r$ is $0.00$), and the top-affected examples have limited overlap (e.g., on MMLU-Pro, the mean Jaccard index is $0.09$ for both the $10\%$ most improved and $10\%$ most degraded examples).
Consistent with this model specificity, question ambiguity is poorly correlated with performance change. By contrast, a model's uncertainty on each question (without context) is meaningfully associated with the magnitude of the corresponding performance change, but not its direction.
Furthermore, we show that context-induced instability persists across a range of context types, tends to increase with context length, can be partially mitigated by scaling test-time compute, and varies substantially across model development stages.

Together, these results identify task-irrelevant context as a concrete source of hidden reliability risk \citep{rabanser2026scienceaiagentreliability} in context-rich LLM deployments.
Example-level prediction flips occur in both directions, with improvements and degradations largely canceling out at the aggregate level.
Moreover, the affected examples are highly model-specific, making it difficult to anticipate which examples will fail based on the data alone. We therefore argue that context robustness evaluation, including per-example instability and worst-tail degradation, should be incorporated into model development to support more trustworthy downstream applications.

\begin{figure}[!t]
  \centering
  \includegraphics[width=0.995\textwidth]{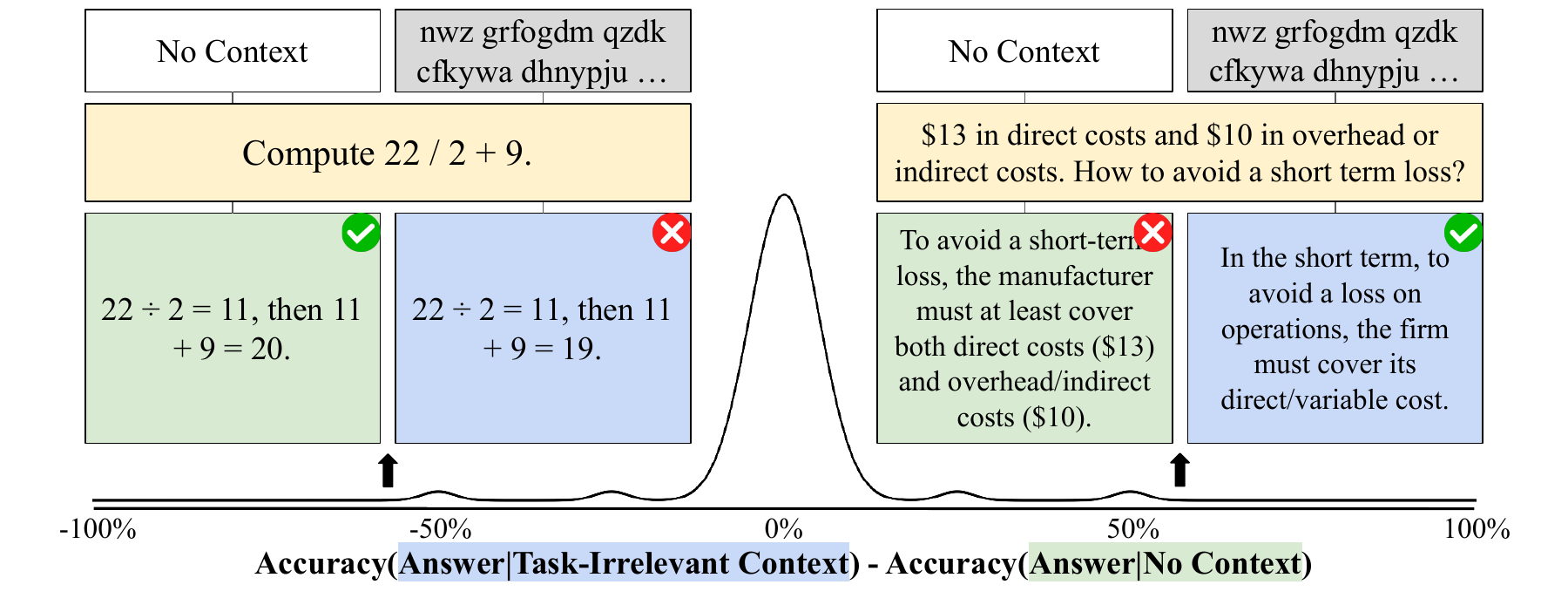}
  \caption{
  \setlength{\fboxsep}{0.8pt}
  \textbf{Context-induced instability is a two-sided effect.} We show two MMLU-Pro examples where task-irrelevant context respectively improves and degrades \texttt{gpt-5.4}'s responses by affecting commonsense and calculation. Full examples are in Appendix~\ref{app: examples}.
  }
  \label{fig:main}
\end{figure}

\section{Setup}
\label{sec: setup}

In our main setup, we use pseudo-words to create task-irrelevant contexts.
Similar to \citep{mohri2026bitterlessondatafiltering}, we first generate a fixed vocabulary of random lowercase strings over the English alphabet (a-z), with lengths between 3 and 8 characters, and filter out strings whose English frequency exceeds a threshold.
For each context, we sampled pseudo-words uniformly from this vocabulary and concatenated them with spaces.
This procedure produces contexts that resemble word-like text at the surface level while minimizing recognizable lexical or factual content.
Note that pseudo-word contexts are not intended to be fully naturalistic. Rather, they provide the cleanest way to isolate the effect.
More naturalistic alternatives, such as randomly sampled tokens, webpages, and question-answer pairs, are tested in our ablation study (Section \ref{subsection: ablation study}).

Let \(D=\{d_1,\ldots,d_N\}\) be the target evaluation dataset, where each example \(d_i\) has reference answer \(y_i\), and let \(C=\{c_1,\ldots,c_M\}\) be a set of task-irrelevant contexts. For a model \(f_{\mathrm{LM}}\), we define $s_{ij}=\Pr(A_{ij}=1 \mid d_i,c_j,f_{\mathrm{LM}})$ as the latent correctness probability of the model on example \(d_i\) when conditioned on context \(c_j\), where \(A_{ij}\in\{0,1\}\) indicates whether the model's response is judged correct. We use \(s_{i0}\) for the corresponding no-context setting.

For a given example $d_i$, the average context-induced performance change is $\Delta_i=\frac{1}{M}\sum_{j=1}^{M}s_{ij}-s_{i0}$, which lies in $[-1, 1]$.
$\Delta_i<0$ indicates that irrelevant context hurts performance on example $d_i$, while $\Delta_i>0$ indicates that irrelevant context improves performance. Across all examples, the average performance change is $\Delta = \frac{1}{N}\sum_{i=1}^{N}\Delta_i$.

From a reliability perspective, a desirable model would show little change under contexts that contain no task-relevant evidence.
Thus, we define the \textbf{Instability (INS)} as the average absolute change:
\[
\mathcal{I}
=
\frac{1}{N}\sum_{i=1}^{N}|\Delta_i|,
\]
where $\mathcal{I} \in [0,1]$.
While INS captures the two-sided sensitivity of model predictions, the more concerning case is when a model that would otherwise answer correctly becomes incorrect after conditioning on irrelevant context.
To isolate the harmful effect, we define \textbf{Worst-tail Degradation (WTD)} as the average degradation in the worst $K\%$ of examples. Let $m=\left\lceil \frac{K}{100}N \right\rceil$, and let $\Delta_{(1)}\leq \Delta_{(2)}\leq \cdots \leq \Delta_{(N)}$ denote the sorted values of $\{\Delta_i\}_{i=1}^N$ in ascending order. We define:
\[
\mathcal{W}_K
=
-\frac{1}{m}\sum_{\ell=1}^{m}\Delta_{(\ell)},
\]
where $\mathcal{W}_K \in [-1,1]$ and a larger $\mathcal{W}_K$ indicates more severe degradation.

\noindent
\textbf{Score Estimation}
Since \(s_{ij}\) is not directly observed, we estimate it in different ways depending on the model access and question format. For API-access models (our main setting), we estimate \(s_{ij}\) by averaging binary correctness over repeated sampled responses. For local models with multiple-choice questions (Appendix \ref{app: local model experiments}), an easier way to estimate it is to use the probability assigned to the correct answer option under a fixed answer template. In both cases, the resulting estimate \(\widehat{s}_{ij}\) is treated as a measurement of the same underlying latent correctness probability.

\noindent
\textbf{Noise Correction}
While calculating $\widehat{\mathcal{I}}$ and $\widehat{\mathcal{W}}_K$ from \(\widehat{s}_{ij}\) for API-access models, finite sampling can induce a positive noise floor for both metrics even when irrelevant context has no true effect, since INS uses absolute values and WTD selects the worst \(K\%\) tail.
To estimate this sampling-induced noise, we use a bootstrap procedure based only on the no-context samples. 
For each example \(d_i\), each bootstrap trial constructs a synthetic null experiment in which all conditions are sampled from the same no-context distribution. Specifically, we independently resample one pseudo-baseline group and \(M\) pseudo-context groups from the same set.
Using the same definitions, each bootstrap trial gives $\widehat{\mathcal{I}}^{*}$ and $\widehat{\mathcal{W}}_{K}^{*}$.
We average over bootstrap trials to get estimated noise floors $\eta_{\mathcal{I}}$ and $\eta_{\mathcal{W}_K}$, and report noise-adjusted metrics by subtracting them from the observed metrics: $\mathcal{I}_{\mathrm{adj}}=\widehat{\mathcal{I}} - \eta_{\mathcal{I}}$ and $\mathcal{W}_{K,\mathrm{adj}}=\widehat{\mathcal{W}}_{K} -\eta_{\mathcal{W}_K}$.
Note that this correction targets the mean null noise floor. For uncertainty, we consider the bootstrap variance of the observed metrics $\widehat{\mathcal{I}}$ and $\widehat{\mathcal{W}}_K$.
Our adjustment should be interpreted as a noise-floor correction rather than an unbiased estimator of the context effect.
In Appendix \ref{app:split-tail}, we provide a complementary split-sample analysis of WTD, where separate samples are used to rank examples and to estimate \(\Delta_i\) within the selected tail.

\begin{table}[t]
\small
\centering
\setlength{\tabcolsep}{3pt}
\renewcommand{\arraystretch}{1.2}
\resizebox{\linewidth}{!}{%
\begin{tabular}{l|rcrr|rcrr|rcrr|rcrr}
\toprule
\toprule
                      & \multicolumn{4}{c|}{\textbf{MMLU-Pro}}     & \multicolumn{4}{c|}{\textbf{GPQA}} & \multicolumn{4}{c|}{\textbf{HLE}}          & \multicolumn{4}{c}{\textbf{SimpleQA}} \\
                                 & AVG   &$\Delta$& INS\textsubscript{$\downarrow$}   & WTD\textsubscript{$\downarrow$} & AVG   &$\Delta$& INS\textsubscript{$\downarrow$}   & WTD\textsubscript{$\downarrow$}   & AVG   &$\Delta$& INS\textsubscript{$\downarrow$}   & WTD\textsubscript{$\downarrow$}   & AVG   &$\Delta$& INS\textsubscript{$\downarrow$}  & WTD\textsubscript{$\downarrow$} \\ \midrule
\texttt{gpt-5.5} & $80.2$ & $+0.3$ & $2.4$ & $10.0$ & $63.8$ & $+0.6$ & $4.9$ & $12.2$ & $12.9$ & $+0.0$ & $1.4$ & $4.3$ & $50.0$ & $-0.9$ & $2.4$ & $11.0$ \\
\texttt{gpt-5.4} & $81.2$ & $-2.1$ & $4.6$ & $26.2$ & $63.1$ & $-0.8$ & $6.2$ & $21.7$ & $11.5$ & $-1.5$ & $3.1$ & $18.8$ & $31.0$ & $+0.4$ & $2.3$ & $7.4$ \\
\texttt{gpt-4.1} & $78.0$ & $+0.0$ & $4.3$ & $19.7$ & $57.3$ & $-0.9$ & $5.6$ & $22.7$ & $3.8$ & $-0.2$ & $1.0$ & $5.4$ & $30.4$ & $-0.7$ & $3.1$ & $11.8$ \\
\texttt{DeepSeek-V4-Pro} & $79.0$ & $+0.2$ & $4.8$ & $15.9$ & $70.0$ & $-0.2$ & $3.0$ & $7.7$ & $7.8$ & $-0.2$ & $1.2$ & $4.2$ & $41.8$ & $-5.9$ & $6.9$ & $27.4$ \\
\texttt{Grok-4.20-NR} & $67.9$ & $+0.1$ & $5.4$ & $21.4$ & $54.5$ & $+0.6$ & $6.6$ & $22.7$ & $9.1$ & $-1.1$ & $2.5$ & $15.6$ & $19.5$ & $-0.4$ & $4.7$ & $20.7$ \\
\texttt{Mistral-Large-3} & $74.9$ & $-3.6$ & $8.3$ & $47.3$ & $50.7$ & $-3.0$ & $11.3$ & $53.2$ & $5.1$ & $+0.0$ & $2.7$ & $12.2$ & $28.7$ & $-1.8$ & $9.6$ & $51.6$ \\ \midrule
\texttt{gpt-oss-120b} & $77.4$ & $-0.6$ & $3.6$ & $15.3$ & $63.4$ & $-2.1$ & $1.2$ & $8.5$ & $4.7$ & $-0.2$ & $0.8$ & $3.3$ & $11.1$ & $-0.5$ & $1.8$ & $9.2$ \\
\texttt{gemini-3.1-fl} & $81.7$ & $+2.3$ & $5.9$ & $17.6$ & $57.2$ & $+8.5$ & $13.6$ & $21.5$ & $6.5$ & $+0.7$ & $3.1$ & $10.6$ & $34.5$ & $-2.0$ & $7.0$ & $34.4$ \\ \bottomrule         
\end{tabular}
}
\vspace{0.1in}
\caption{
  \setlength{\fboxsep}{0.8pt}
  \textbf{Main results ($\%$).} AVG refers to the average baseline accuracy, $\Delta$ refers to the average performance change after adding contexts, and INS and WTD are noise-floor adjusted. Models on the top use no reasoning by default, while models on the bottom use the lowest reasoning effort.
  }
  \label{table:main}
\end{table}

\begin{figure}[!t]
  \centering
  \includegraphics[width=\textwidth]{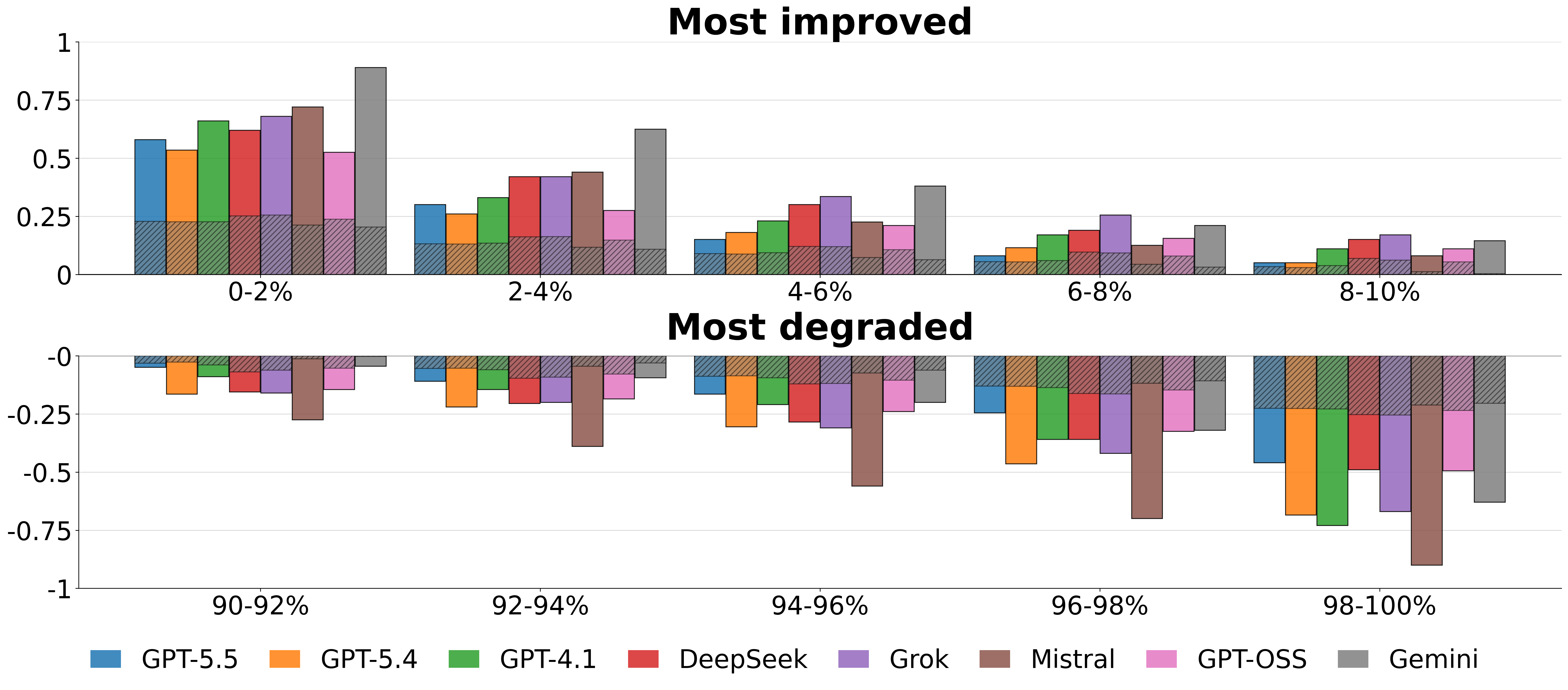}
  \caption{
  \setlength{\fboxsep}{0.8pt}
  \textbf{Per-example performance change concentrates in the tails (MMLU-Pro).} For each model, we rank its examples from most improved to most degraded by the per-example change \(\Delta_i\), and keep the two \(10\%\) tails: the most-improved examples (top panel) and the most-degraded examples (bottom panel). Each tail is split into \(2\%\)-wide percentile bins (x-axis), and each bar reports the mean \(\Delta_i\) in that bin for one model. Solid bars show the observed change; the overlaid hatched bars show the corresponding estimated noise floor.
  }
  \label{fig:distribution}
\end{figure}

\begin{table}[t]
  \centering
  \setlength{\tabcolsep}{4pt}
  \begin{tabular}{l|rr|rr|rr|rr}
  \toprule
  \toprule
   & \multicolumn{2}{c|}{\textbf{MMLU-Pro}} & \multicolumn{2}{c|}{\textbf{GPQA}} & \multicolumn{2}{c|}{\textbf{HLE}} & \multicolumn{2}{c}{\textbf{SimpleQA}} \\ \midrule
   & Impr. & Degr. & Impr. & Degr. & Impr. & Degr. & Impr. & Degr. \\
  \midrule
  Wrong Problem     & $21.1$ & $6.3$  & $18.9$ & $4.8$  & $31.4$ & $19.4$ & $2.1$  & $0.8$  \\
  Wrong Fact        & $21.1$ & $7.4$  & $24.4$ & $26.7$ & $12.9$ & $34.4$ & $97.9$ & $96.0$ \\
  Wrong Rule        & $21.1$ & $24.7$ & $33.3$ & $25.7$ & $42.9$ & $38.8$ & $0.0$  & $0.0$  \\
  Wrong Calculation & $36.8$ & $61.6$ & $23.3$ & $42.9$ & $12.9$ & $6.2$  & $0.0$  & $0.0$  \\
  Other or unclear  & $0.0$  & $0.0$  & $0.0$  & $0.0$  & $0.0$  & $1.2$  & $0.0$  & $3.2$  \\
  \bottomrule
  \end{tabular}
  \vspace{0.1in}
  \caption{\label{table: error distribution}
    \textbf{Qualitative analysis of performance changes.} For questions whose accuracy changes by more than $20\%$ after adding task-irrelevant context, we annotate the behavioral differences between the no-context and with-context settings using five pairwise comparisons per question. The table reports the distribution ($\%$) of annotated error types across pairs. For improved questions, the categories indicate the errors corrected by the added context; for degraded questions, they indicate the errors introduced.
  }
\end{table}

\section{Experiments}

\noindent
\textbf{Experiment Details}
We use the following four datasets: MMLU-Pro \citep{wang2024mmluprorobustchallengingmultitask}, GPQA-Diamond \citep{rein2023gpqagraduatelevelgoogleproofqa}, Humanity's Last Exam \citep{phan2025humanity}, and SimpleQA Verified \citep{wei2024measuringshortformfactualitylarge, haas2026simpleqaverifiedreliablefactuality}.
The first two datasets contain multiple-choice questions, while the latter two contain open-ended questions.
For each dataset, we randomly sample up to $500$ questions.
We create $M=10$ pseudo-word contexts, each with $10k$ tokens by default ($21129.9$ characters on average).
For consistency, we use token counts from the \texttt{gpt-oss-120b} tokenizer to create contexts in this work.
Contexts are prepended to the question as an extra user message between the question and the system message.
We use $K=10$ for the WTD metric.
By default, we use API-access models and report the adjusted version of INS and WTD, while providing extra results for local models in the ablation study and Appendix \ref{app: local model experiments}.
For each question in the experiments, we use a temperature of $1.0$ to sample $p=20$ times for the no-context baseline and $q=2$ times for each context, so that both conditions have 20 trials in total.
We report results for \texttt{gpt-5.5}, \texttt{gpt-5.4}, \texttt{gpt-4.1}, \texttt{DeepSeek-V4-Pro}, \texttt{Grok-4.20-NR} (\texttt{Grok 4.20 non reasoning}), \texttt{Mistral-Large-3}, \texttt{gpt-oss-120b}, and \texttt{gemini-3.1-fl} (\texttt{gemini-3.1-flash-lite}).
By default, we use no reasoning setup to rule out the effect of reasoning tokens.
For models that cannot turn off reasoning, we use the lowest possible reasoning effort (\texttt{low} or \texttt{minimal}).
We use \texttt{gpt-5.4-mini} with \texttt{low} reasoning effort to grade all responses.
Extra experiment details are in Appendix \ref{app: extra exp details} and prompt details are in Appendix \ref{app: prompt}.

\subsection{Main results}
We show the main result in Table~\ref{table:main} (unadjusted metrics, confidence intervals, and noise floors are shown in Appendix Table~\ref{tab:ins-unr-ci}).
With the added pseudo-word contexts, the average performance change $\Delta$ usually remains close to zero across models and benchmarks.
However, the INS metrics are usually much larger (reaching $13.6\%$), suggesting that the aggregated stability is not due to per-example stability but rather to two-sided effects that cancel out.
If we take a closer look at the per-example distribution  (Figure \ref{fig:distribution}), we find that the instability mostly comes from the top $10\%$ tail on both sides, accounting for $88.3\%$ of the total absolute change.
For the $10\%$ examples most negatively impacted, WTD indicates that the average performance degradation can be as large as $53.2\%$.

Among the evaluated models without reasoning, \texttt{gpt-5.5} is the most consistently stable across all four benchmarks. Its INS is at most $4.9\%$, and its WTD is at most $12.2\%$.
Note that low INS or WTD should be interpreted together with the baseline accuracy.
For example, \texttt{gpt-4.1} has low INS and WTD on HLE, but its baseline AVG is only $3.8\%$.
In this case, there is limited room for performance changes, so small INS can partly reflect a floor effect rather than genuine robustness.
More importantly, it is possible to achieve both strong baseline accuracy and low context sensitivity.
On SimpleQA, for instance, \texttt{gpt-5.5} reaches $50.0\%$ AVG with only $2.4\%$ INS and $11.0\%$ WTD, whereas several models with substantially lower AVG exhibit larger INS and WTD.

\noindent
\textbf{Improvement vs. Degradation}
Models can have similar INS but very different directional behavior.
For example, both \texttt{Mistral-Large-3} and \texttt{gemini-3.1-fl} exhibit large INS across several settings, but their instability manifests differently.
\texttt{Mistral-Large-3} tends to exhibit degradation, leading to particularly high WTD values: $47.3\%$ on MMLU-Pro, $53.2\%$ on GPQA, and $51.6\%$ on SimpleQA.
In contrast, \texttt{gemini-3.1-fl} often shows relatively positive shifts, especially on MMLU-Pro and GPQA, where $\Delta$ is $+2.3\%$ and $+8.5\%$, respectively.
In Figure \ref{fig:distribution}, \texttt{gemini-3.1-fl} has larger bars in the improved tail, while \texttt{Mistral-Large-3} is more prominent in the degraded tail.
Thus, high INS alone does not indicate whether the irrelevant context is mostly helpful or harmful; WTD is necessary to isolate the harmful tail.

\noindent
\textbf{Qualitative Analysis}
To better understand context-induced instability, we use LLM-based annotations to analyze questions whose accuracy changes by more than $20\%$ after adding task-irrelevant context for \texttt{gpt-5.4} (details in Appendix \ref{app: qualitative analysis}).
Table \ref{table: error distribution} summarizes the types of errors that are corrected in improved examples and introduced in degraded examples.
Overall, the dominant error types reflect the nature of each benchmark. On MMLU-Pro, performance changes are primarily associated with calculation errors, including arithmetic mistakes, bookkeeping errors, and miscounts. On HLE, changes are more often linked to the application of incorrect or overgeneralized rules, principles, or concepts. On SimpleQA, nearly all changes correspond to factual hallucinations, consistent with its fact-seeking format.
Importantly, improved and degraded examples do not follow identical error distributions. In MMLU-Pro and GPQA, degradation is disproportionately associated with newly introduced calculation errors. In contrast, solving the wrong problem is more common among improved examples in MMLU-Pro, GPQA, and HLE.
These results suggest that irrelevant context can alter the model’s predictions in qualitatively different ways.

\begin{figure}[t]
  \centering
  \includegraphics[width=\textwidth]{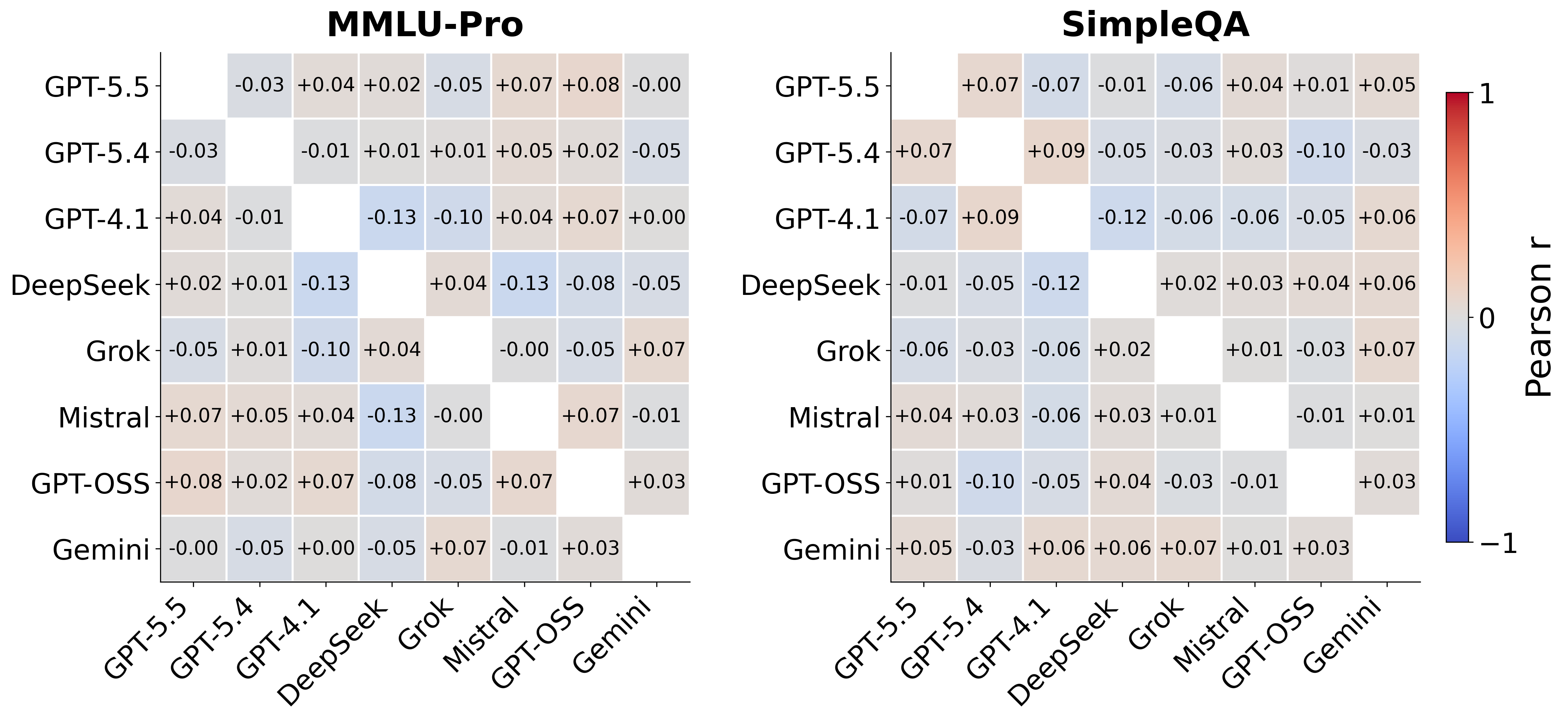}
  \caption{
  \setlength{\fboxsep}{0.8pt}
  \textbf{Per-example changes are uncorrelated across models.} We report Pearson $r$ of per-example performance change for all model pairs on MMLU-Pro and SimpleQA.
  }
  \label{fig:correlation}
\end{figure}

\subsection{Model Specificity}

When we examine per-example changes $[\Delta_1, \Delta_2, \cdots, \Delta_N]$, there is almost no correlation across models, even among those with similar AVG/INS/WTD in Table \ref{table:main}.
On all four datasets, we plot the Pearson correlation between all models in Figure \ref{fig:correlation} (and Appendix Figure \ref{fig:correlation app}). The mean Pearson correlation is $0.00$ ($95\%$ CI: $[-0.02, 0.02]$) on MMLU-Pro and similarly $0.00$ ($95\%$ CI: $[-0.02, 0.01]$) on SimpleQA.
If we take a closer look at the correlations between model pairs, there is no noticeable positive correlation between any two models across these two datasets.
We also calculate the Jaccard overlap of the top $10\%$ improved/degraded examples. On average, the Jaccard overlap is $0.09$ ($95\%$ CI: $[0.08, 0.11]$) for improved examples and $0.09$ ($95\%$ CI: $[0.08, 0.11]$) for degraded examples on MMLU-Pro, which is close to the $0.05$ random baseline.

\begin{table*}[t]
\small
\centering
\setlength{\tabcolsep}{3pt}
\renewcommand{\arraystretch}{1.2}
\resizebox{\linewidth}{!}{%
  \begin{tabular}{llcccc}
    \toprule
    \toprule
    & & \multicolumn{2}{c}{Raw} & \multicolumn{2}{c}{Split-sample} \\
    \cmidrule(lr){3-4}\cmidrule(lr){5-6}
  Feature & Target & Probability & Logit & Probability & Logit \\
  \midrule
  Ambiguity   & Signed  & $-0.01$ {\scriptsize $[-0.06,\,+0.04]$} & $-0.01$ {\scriptsize $[-0.05,\,+0.04]$} & $-0.01$ {\scriptsize $[-0.05,\,+0.03]$} & $-0.01$ {\scriptsize $[-0.05,\,+0.04]$} \\
              & Squared & $+0.02$ {\scriptsize $[-0.03,\,+0.06]$} & $+0.02$ {\scriptsize $[-0.03,\,+0.06]$} & $+0.02$ {\scriptsize $[-0.02,\,+0.05]$} & $+0.02$ {\scriptsize $[-0.02,\,+0.05]$} \\
  \addlinespace
  Uncertainty & Signed  & $-0.08$ {\scriptsize $[-0.16,\,-0.00]$} & $-0.08$ {\scriptsize $[-0.15,\,-0.01]$} & $-0.03$ {\scriptsize $[-0.08,\,+0.03]$} & $-0.02$ {\scriptsize $[-0.07,\,+0.04]$} \\
              & Squared & $+0.73$ {\scriptsize $[+0.70,\,+0.76]$} & $+0.66$ {\scriptsize $[+0.62,\,+0.70]$} & $+0.21$ {\scriptsize $[+0.17,\,+0.26]$} & $+0.21$ {\scriptsize $[+0.16,\,+0.25]$} \\
    \bottomrule
  \end{tabular}
  }
  \caption{
    \textbf{Correlates of context-induced performance change.}
    Spearman $\rho$ with 95\% question-bootstrap CIs over all $n{=}1690$ questions. We correlate question ambiguity and baseline uncertainty with the signed change $\Delta_i$ (direction) and squared change $\Delta_i^2$ (magnitude). Raw estimates use all trials, while split-sample estimates use disjoint trials for the feature and target. We report both probability-scale and smoothed log-odds-scale results. Estimator details are in Appendix~\ref{app:split-sample-correlation}.
    \label{tab:quality-null}}
\end{table*}

\noindent
\textbf{Correlates of Context-Induced Change}
For each question, we measure ambiguity by asking three judge models to flag ambiguity or underspecification and summing their positive flags into a score in $\{0,1,2,3\}$ (Appendix~\ref{app: ambiguity}).
We use the estimated Bernoulli variance of no-context correctness, $\widehat{s}_{i0}(1-\widehat{s}_{i0})$, as a proxy for baseline predictive uncertainty.
Table~\ref{tab:quality-null} correlates these features with the signed change $\Delta_i$, which captures whether context helps or hurts, and the squared change $\Delta_i^2$, which captures its magnitude. We report raw estimates together with split-sample estimates that reduce shared-trial noise on both the probability and log-odds scales.
Ambiguity is consistently uncorrelated with either the direction or magnitude of change. Baseline uncertainty is also nearly uncorrelated with direction. Its association with magnitude is strong in the raw estimates and remains modest after split-sampling ($\rho{=}0.21$). Importantly, the same association remains on the log-odds scale, suggesting that it is not explained solely by uncertain questions having more room to change on the probability scale. Intuitively, a model's internal uncertainty helps identify questions that are more likely to change, but not whether the change will be beneficial or harmful.

\begin{figure}[t]
  \centering
  \includegraphics[width=\textwidth]{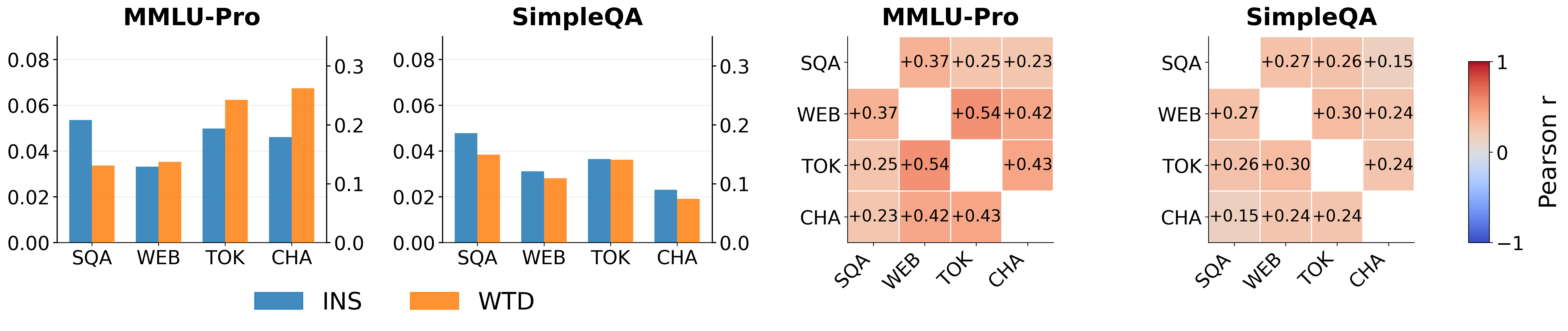}
  \caption{
  \setlength{\fboxsep}{0.8pt}
  \textbf{Context-induced instability holds across context types.} We show INS, WTD, and the correlation of per-example performance change between different context types. SQA refers to sequential question answering, WEB refers to random webpages, TOK refers to random tokens, and CHA refers to pseudo-words. All results are based on \texttt{gpt-5.4} and MMLU-Pro.
  }
  \label{fig:instability}
\end{figure}

\begin{figure}[t]
  \centering
  \includegraphics[width=\textwidth]{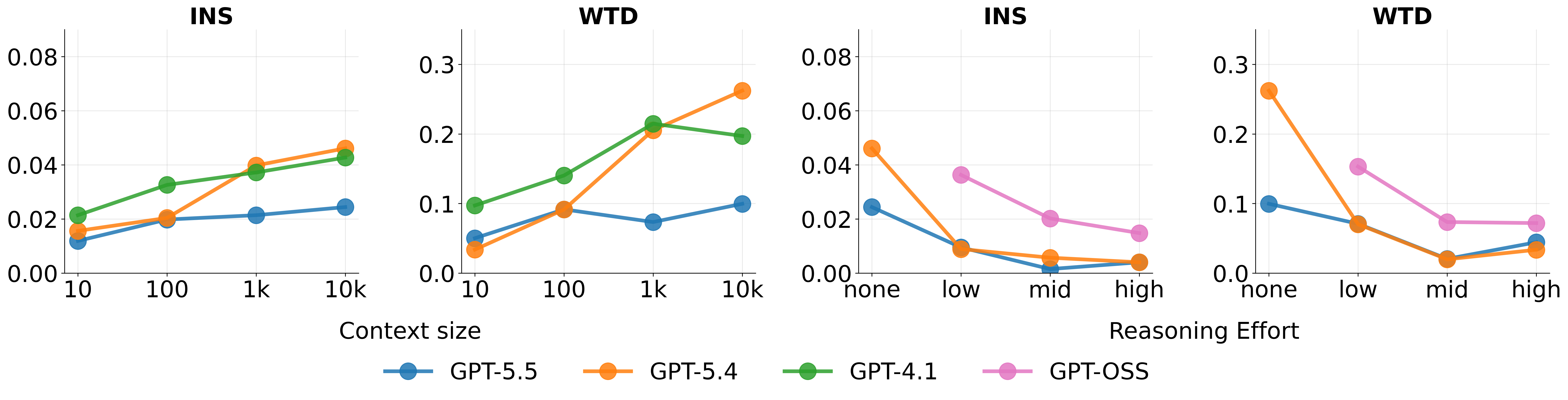}
  \caption{
  \setlength{\fboxsep}{0.8pt}
  \textbf{Scaling effects of context length and reasoning effort.} We vary context length ($10$, $100$, $1k$, and $10k$) and reasoning effort (\texttt{none}, \texttt{low}, \texttt{medium}, \texttt{high}) on MMLU-Pro.
  }
  \label{fig:context and reasoning}
\end{figure}

\subsection{Ablation Study}
\label{subsection: ablation study}

\noindent \textbf{Context Type}
Beyond pseudo-word contexts constructed by random characters (CHA), we further evaluate three more naturalistic types of task-irrelevant context. First, \textbf{Random Tokens} (TOK) are constructed by randomly sampling tokens from the \texttt{gpt-oss-120b} tokenizer vocabulary. Second, \textbf{Random Webpages} (WEB) are formed by randomly sampling and concatenating webpages from \texttt{HuggingFaceFW/fineweb-edu}. Third, \textbf{Sequential Question Answering} (SQA) consists of multi-turn conversation histories in which the model answers other randomly sampled questions from the same dataset. For TOK and WEB, each context contains $10k$ tokens; for SQA, we use $20$ turns by default.
Unlike pseudo-word contexts, these settings contain real words, webpages, or questions, and may therefore include superficially or semantically related information by chance. We use them as a sanity check to test whether context-induced instability also appears under more naturalistic forms of task-irrelevant context, and to measure whether the affected questions are consistent.

As shown in Figure~\ref{fig:instability}, all context types induce both INS and WTD, although the magnitudes vary across datasets and context types. In both datasets, SQA yields the highest INS, likely because its question-answer format most closely resembles the target task and therefore more strongly perturbs the model's prediction. The default pseudo-word context yields the highest WTD on MMLU-Pro but the lowest WTD on SimpleQA, suggesting that the same context type can have different relative effects depending on the benchmark. Despite these differences in magnitude, the question-level effects are positively correlated across context types, with mean Pearson correlations of $0.37$ ($95\%$ CI: $[0.28, 0.46]$) on MMLU-Pro and $0.24$ ($95\%$ CI: $[0.16, 0.33]$) on SimpleQA. This indicates that while the magnitude of instability depends on the form of irrelevant context, the questions affected by different context types partially overlap. We report results with local models in Appendix \ref{app: local model experiments}.

\noindent \textbf{Context Length}
Figure~\ref{fig:context and reasoning} (Left) examines how context length affects INS and WTD using pseudo-word contexts of $10$, $100$, $1k$, and $10k$ tokens. Surprisingly, even very short contexts containing only $10$ pseudo-word tokens induce nontrivial INS and WTD. Increasing the context length generally amplifies both effects, although the trends saturate at different lengths for different models. For example, the WTD of \texttt{gpt-5.4} continues to increase as the context length grows from $10$ to $10k$ tokens, whereas the WTD of \texttt{gpt-5.5} saturates after $100$ context tokens.
We further find that the pattern of per-example changes evolves as context length increases. For instance, the Pearson correlation between the changes induced by contexts of length $10$ and $100$ is $r=0.59$ for \texttt{gpt-5.4} and $r=0.54$ for \texttt{gpt-5.5}. In contrast, the correlation between contexts of length $10$ and $10k$ decreases to $r=0.08$ for \texttt{gpt-5.4} and $r=0.24$ for \texttt{gpt-5.5}. These results suggest that increasing context length affects not only the magnitude of the impact on individual questions, but also which questions are affected.

\noindent \textbf{Test-time Compute}
First, we examine the effect of increasing reasoning effort for reasoning models in Figure~\ref{fig:context and reasoning} (Right) with details in Appendix Table \ref{tab:reasoning-misc}. For \texttt{gpt-5.4}, increasing reasoning effort from \texttt{none} to \texttt{low} sharply reduces both INS and WTD, largely closing the gap with \texttt{gpt-5.5}. However, additional reasoning does not monotonically reduce WTD. For example, increasing reasoning effort from \texttt{medium} to \texttt{high} does not decrease WTD for either \texttt{gpt-5.4} or \texttt{gpt-5.5}. As a result, it does not fully eliminate the issue, and significantly degraded examples still exist. A qualitative inspection of \texttt{gpt-5.4} with \texttt{high} reasoning effort shows that wrong calculation errors are largely eliminated, but solving the wrong problem and applying the wrong rule remain common failure modes.
Second, we apply Chain-of-Thought (CoT) prompting \citep{wei2022chain} for a non-reasoning model \texttt{gpt-4.1} (Full results in Appendix Table \ref{table:ablation 4.1}). Similar to increasing reasoning effort, CoT prompting also reduces both INS and WTD across all datasets, although the degree of mitigation varies substantially: INS decreases from $4.3\%$ to $1.3\%$ and WTD from $19.7\%$ to $6.4\%$ on MMLU-Pro. On SimpleQA, INS decreases from $3.1\%$ to $1.7\%$ and WTD from $11.8\%$ to $7.6\%$. The remaining WTD is still relatively high on SimpleQA, suggesting that factual errors may be harder to recover even with a more explicit thinking process.
Overall, these results indicate that spending additional tokens at test time can mitigate context-induced instability, but it does not provide a complete solution.

\begin{figure}[t]
  \centering
  \includegraphics[width=\textwidth]{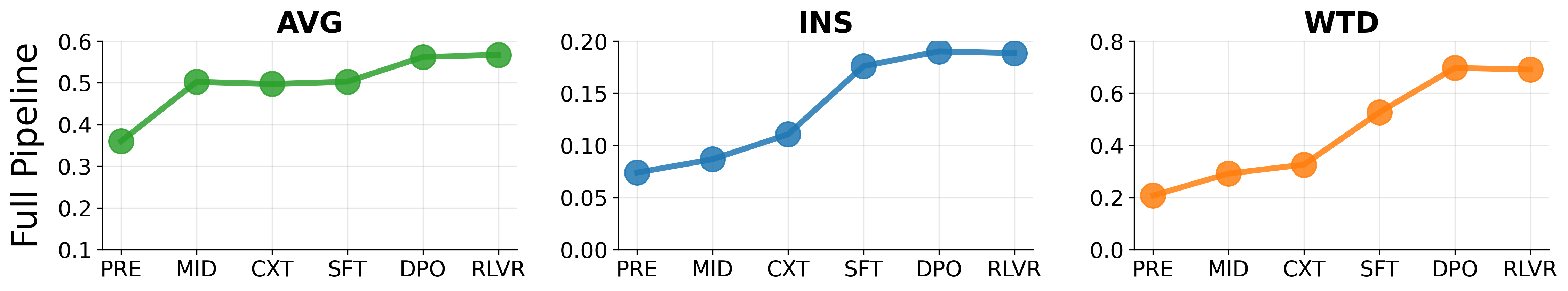}
  \includegraphics[width=\textwidth]{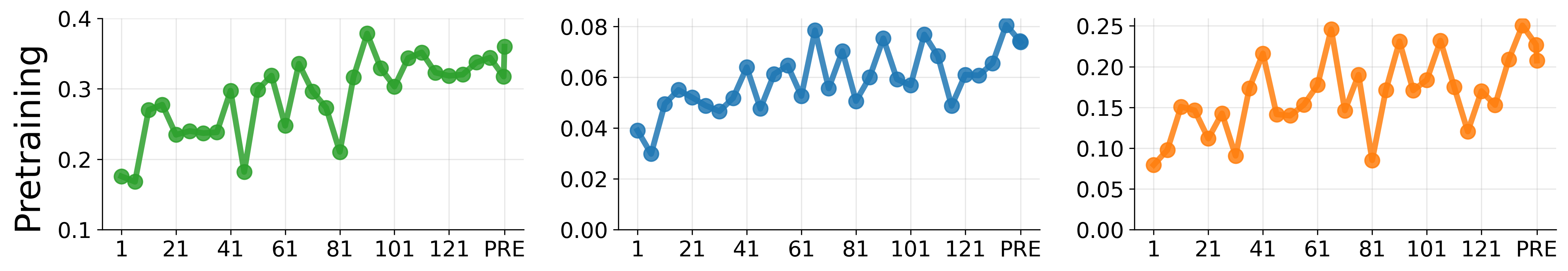}
  \includegraphics[width=\textwidth]{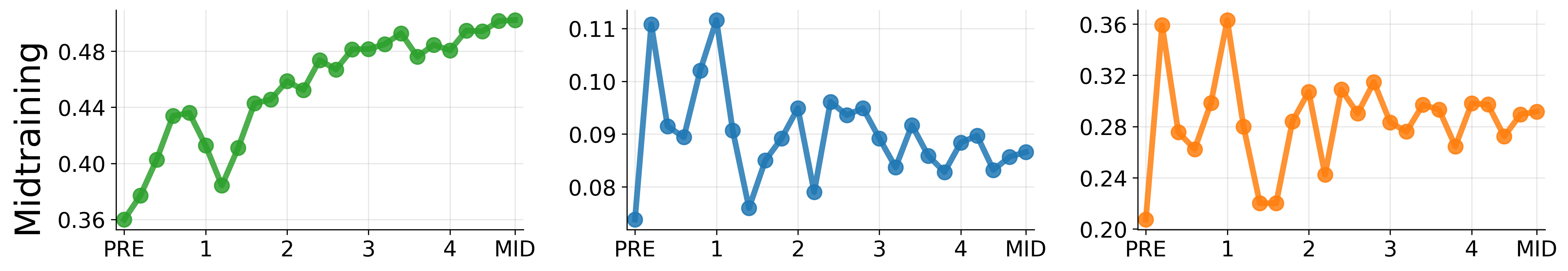}
  \caption{
  \setlength{\fboxsep}{0.8pt}
  \textbf{Context-induced instability emerges across the training pipeline.} We track INS and WTD on MMLU for \texttt{Olmo-3-7B-Instruct} at checkpoints (I) after each training stage, (II) during pretraining, and (III) during midtraining.
  }
  \label{fig:training dynamic}
\end{figure}

\begin{table}[t]
\small
\centering
\setlength{\tabcolsep}{3pt}
\renewcommand{\arraystretch}{1.2}
{
\newcommand{\incinst}[1]{\,{\scriptsize\textcolor{red!70!black}{($+#1$)}}}
\newcommand{\decinst}[1]{\,{\scriptsize\textcolor{green!70!black}{($-#1$)}}}
\resizebox{\linewidth}{!}{%
\begin{tabular}{l|rcll|l|rcll}
\toprule
\toprule
           & AVG & $\Delta$ & INS\textsubscript{$\downarrow$}   & WTD\textsubscript{$\downarrow$} &
           & AVG & $\Delta$ & INS\textsubscript{$\downarrow$}   & WTD\textsubscript{$\downarrow$} \\
\midrule
\texttt{Qwen3.5-9B-Base}     & $72.5$ & $-1.0$ & $3.7$  & $13.7$ &
\texttt{Olmo-3-7B-2T}        & $25.4$ & $-2.1$ & $5.0$  & $17.7$ \\
\texttt{Qwen3.5-9B}          & $73.8$ & $-4.5$ & $6.3$\incinst{2.6}  & $21.3$\incinst{7.6} &
\texttt{2T + Gen-QA}        & $44.9$ & $-1.6$ & $6.5$\incinst{1.5}  & $20.0$\incinst{2.3} \\
\texttt{gemma-4-12B}         & $58.5$ & $-0.5$ & $6.7$  & $24.5$ &
\texttt{2T + Math-code}
                                & $43.1$ & $-1.3$ & $5.2$\incinst{0.2}  & $16.1$\decinst{1.6} \\
\texttt{gemma-4-12B-it}      & $78.5$ & $-0.1$ & $8.4$\incinst{1.7}  & $40.7$\incinst{16.2} &
\texttt{2T + Final}         & $47.4$ & $-2.1$ & $6.9$\incinst{1.9}  & $22.1$\incinst{4.4} \\
\bottomrule
\end{tabular}}}
\vspace{0.1in}
\caption{
\textbf{INS and WTD for local models on MMLU.} Left: comparison between base and post-trained models. Right: OLMo-3 checkpoints with different midtraining data mixtures.
}
\label{table:training-dynamic}
\end{table}

\noindent \textbf{Model Development Stage}
To understand how context-induced instability evolves during training, we evaluate local models on the original MMLU benchmark~\citep{hendrycks2021measuringmassivemultitasklanguage}, focusing on the OLMo-3 family~\citep{olmo2026olmo3}, which releases intermediate checkpoints throughout training. Figure~\ref{fig:training dynamic} tracks no-context performance (AVG), INS, and WTD across the development stages of \texttt{Olmo-3-7B-Instruct}: pretraining (PRE), midtraining (MID), context extension (CXT), supervised finetuning (SFT), direct preference optimization (DPO), and reinforcement learning from verifiable rewards (RLVR).
\textbf{(I)} We find that INS and WTD gradually emerge during pretraining. As task performance improves, the model also becomes increasingly sensitive to task-irrelevant context. During midtraining, both INS and WTD initially increase sharply, possibly reflecting a distribution shift in the training data, before gradually stabilizing at lower values, as task performance continues to improve. Later stages exhibit different patterns. SFT changes baseline performance only modestly, but substantially increases instability. The subsequent DPO stage further improves task performance while also amplifying INS and WTD. We observe a similar post-training effect in Table~\ref{table:training-dynamic}: both \texttt{Qwen3.5-9B} and \texttt{gemma-4-12B-it} exhibit higher INS and WTD than their corresponding base models.
These trends suggest that context-induced instability is shaped throughout all stages. In particular, improvements in benchmark performance do not necessarily imply improved robustness to irrelevant context; in some stages, especially post-training, performance gains coincide with increased instability.
\textbf{(II)} We next examine whether instability is associated with the choice of data mixture for the same stage. We analyze three OLMo-3 midtraining checkpoints that start from the same 2T-token pretrained checkpoint but use different data mixtures. We find that midtraining on Generative/QA-domain data is associated with higher INS and WTD than midtraining on code/math/thinking-domain data, even though both mixtures substantially improve AVG performance. Interestingly, the final OLMo-3 mixture, which more evenly balances Generative/QA and code/math/thinking data, leads to the highest instability among the three. While these comparisons are observational, they suggest that the composition of training data may also influence context-induced instability.

\section{Related work}

\noindent
\textbf{Robustness to Prompt Perturbations}
For classic NLP tasks, CheckList~\citep{ribeiro2020accuracybehavioraltestingnlp} introduces invariance-style tests for perturbations that should not affect model predictions. For LLMs, a natural expectation is that predictions should be stable under prompt wording, formatting, and other semantically preserving variations, but prior work shows that this expectation often fails. \citet{sclar2024quantifyinglanguagemodelssensitivity} show that meaning-preserving prompt-format changes can shift accuracy by up to $76\%$. \citet{mizrahi2024stateartmultipromptllm} argue that single-prompt evaluation is brittle and that models should be evaluated with diverse prompts. \citet{zhuo2024prosaassessingunderstandingprompt} provide instance-level analyses of prompt sensitivity and find that robustness improves with more few-shot examples and higher model confidence. \citet{salinas2024butterflyeffectalteringprompts} further show that even minor formatting changes, such as adding a trailing space, can change an LLM's answer. In this work, the same invariance expectation extends beyond the prompt template to the input itself: a model's per-example performance should not change substantially when task-irrelevant context is added.

\noindent
\textbf{Robustness to Irrelevant Context}
Whereas one line of work asks whether models can retrieve \emph{relevant} evidence from the context, where performance depends strongly on where evidence appears~\citep{liu2023lostmiddlelanguagemodels} and degrades as context length and task complexity increase~\citep{kamradt2023needle,hsieh2024rulerwhatsrealcontext}, a complementary line studies the effect of \emph{irrelevant} context. \citet{shi2023largelanguagemodelseasily} show that irrelevant information can substantially degrade arithmetic reasoning, while \citet{yoran2024makingretrievalaugmentedlanguagemodels} show that retrieval-augmented models can be harmed by irrelevant passages unless explicitly trained to handle them. More recently, \citet{wu2024reducingdistractionlongcontextlanguage} propose focused learning to reduce distraction in long-context QA, \citet{yang2025llmreasoningdistractedirrelevant} introduce a controlled benchmark for distracting context in mathematical reasoning, and \citet{wang2025adaptivedistractionprobingllm} use automated tree search to generate adaptive distractions that strongly reduce performance.
Beyond strategically designed distractors, \citet{levy-etal-2024-task} and \citet{du-etal-2025-context} show that irrelevant filler text and even whitespace can degrade model performance, and \citet{rodionov2026reasoningshiftcontextsilently} recently show that lengthy irrelevant context can shorten the reasoning trace and degrade performance. Conversely, \citet{shi2025meaninglesstokensmeaningfulgains} show that inserting a carefully tuned number of repeated meaningless tokens can improve performance. However, each of these works examines only a limited set of datasets, models, and context types, and thus does not provide a comprehensive picture. In this work, with broad coverage along these axes, we reveal a general phenomenon of context-induced instability: even when aggregate accuracy remains stable, individual examples can undergo large, model-specific performance shifts in either direction.

\noindent
\textbf{Mechanisms of Context Sensitivity}
A growing body of work seeks to explain the context-related behavior of LLMs. \citet{wu2025retrievalhead} identify attention heads responsible for retrieving information from the context, and \citet{zhu2025focusdirections} label relevant context spans and learn ``focus'' directions to mitigate distraction. Closest to our setting, \citet{niu-etal-2025-llama} describe contextual entrainment: models assign higher probability to tokens that previously appeared in the context, regardless of their semantic relevance, and attribute the effect to a mechanistic token-copying bias. \citet{kukreja-etal-2026-better} further report that for non-semantic context, this entrainment increases with model scale. This work instead uses pseudo-word context, which further minimizes lexical overlap between the context and the question or answer. Even so, it still induces performance shifts on state-of-the-art models and challenging benchmarks, suggesting that an overlap-driven copying bias alone does not account for the effect. We hope it offers a new perspective for understanding and improving context robustness.

\section{Conclusion}

In this work, we study context-induced instability in LLMs by adding task-irrelevant context to standalone benchmark questions. Across models and datasets, we find that such context can substantially change per-example behavior, degrading performance on some questions while improving it on others. These affected examples are largely model-specific, making them difficult to predict from the data alone. We further find that context-induced instability evolves across model development stages and can increase alongside benchmark performance, while additional test-time reasoning mitigates but does not eliminate it.
Building on these findings, future work could explore several directions:
\begin{itemize}[leftmargin=1.5em, itemsep=2pt, topsep=2pt]
\item \textbf{Improving measurement efficiency.} Reliably estimating tail metrics such as WTD currently requires repeated sampling over large datasets to ensure that the worst-$K\%$ tail contains enough examples, resulting in substantial computational cost.
\item \textbf{Understanding the sources of instability.} Explaining why task-irrelevant context destabilizes particular examples, from both training-data and mechanistic interpretability perspectives, could make training and inference methods more robust.
\item \textbf{Diagnosing agent failures.} A deeper understanding of end-to-end agent failures could be achieved by more carefully isolating the effects of context and quantifying how instability at individual steps accumulates across multi-step trajectories.
\end{itemize}

\section*{Acknowledgment}
We would like to thank all SALT Lab members and friends for their feedback.
This work is supported by Open Philanthropy, Schmidt Sciences, and a grant under ONR N00014-24-1-2532.

\bibliography{main}
\bibliographystyle{abbrvnat}

\newpage

\appendix

\section{Split-sample WTD}
\label{app:split-tail}

For WTD, the naive estimator ranks questions by $\widehat{\Delta}_i$, keeps the $m=\left\lceil \frac{K}{100}N \right\rceil$ most degraded of the $N$ questions, and reports the negative mean. Evaluating the same noisy estimates used for selection exaggerates degradation. The main paper uses a bootstrap procedure as its default noise-floor correction; here we report a complementary split-sample estimation in Table \ref{tab:split-sample-tail}.

To remove the bias, we select and measure on \emph{disjoint} trials. We split each side into two:
\begin{itemize}
\item The baseline trials are split into two disjoint halves $B_1, B_2$;
\item The context trials are split into two disjoint halves $C_1, C_2$.
\end{itemize}
On each half, we form an independent estimate of the change,
\begin{align}
    d_1 \;=\; \mathrm{mean}(C_1) - \mathrm{mean}(B_1), \qquad
    d_2 \;=\; \mathrm{mean}(C_2) - \mathrm{mean}(B_2),
\end{align}
We use one half to \emph{select} the tail and the other to \emph{measure} it,
\begin{align}
S \;=\; \{\,\text{the } m \text{ questions with the smallest } d_1\,\},
\qquad
\widehat{\mathcal{W}}_{K} \;=\; -\frac{1}{m}\sum_{i \in S} d_{2,i}.
\end{align}
Because the selected set $S$ depends only on $d_1$, and $d_2$ is independent of $d_1$ given the question, \(\mathbb{E}[d_{2,i}\mid i\in S]\) equals the true change of each selected question. Thus, measurement is unbiased for the questions selected by the noisy ranking. The selected set can still differ from the oracle tail ranked by the latent \(\Delta_i\).
A single split is noisy, so we repeat over random splits and average $\widehat{\mathcal{W}}_{K}$. By default, the two halves are also used \emph{symmetrically} (select on $d_1$ / measure $d_2$, and select on $d_2$ / measure $d_1$, then average): every trial feeds both selection and measurement.

\section{Extra Experiment Details}
\label{app: extra exp details}

While building the pseudo-word vocabulary, the Zipf frequency threshold used for the English word filtering is $1.5$, and the total vocabulary size is $10000$.

For each dataset, we first apply keyword-based filtering to remove questions that might trigger abuse policies (e.g., biosafety and cybersecurity). Then we randomly sample 500 examples for MMLU-Pro, HLE (text-only questions), and SimpleQA Verified.
Since GPQA-Diamond originally has only 198 questions and 8 are filtered, we use the remaining 190 questions.

We use the following parameters consistently. To estimate the noise floor, we construct $1000$ synthetic null experiments as described in Section~\ref{sec: setup}. To estimate each $95\%$ confidence interval, we bootstrap questions with replacement $1000$ times, retaining all conditions and trials associated with each sampled question and recomputing the complete reported statistic. In split-sample and split-half experiments, we average over $200$ random trial splits.

\begin{table}[t]
\centering
\small
\setlength{\tabcolsep}{3pt}
\renewcommand{\arraystretch}{1.2}
\resizebox{\linewidth}{!}{
\begin{tabular}{l|c|c|c|c}
\toprule
\toprule
    & \textbf{MMLU-Pro} & \textbf{GPQA} & \textbf{HLE} & \textbf{SimpleQA} \\
\midrule
\texttt{GPT-5.5} & $12.0$ {\scriptsize $[5.8, 18.4]$} & $15.6$ {\scriptsize $[6.9, 24.4]$} & $5.3$ {\scriptsize $[1.9, 9.5]$} & $14.1$ {\scriptsize $[8.5, 19.8]$} \\
\texttt{GPT-5.4} & $29.0$ {\scriptsize $[19.9, 38.1]$} & $26.0$ {\scriptsize $[16.7, 35.4]$} & $22.4$ {\scriptsize $[14.7, 30.2]$} & $10.7$ {\scriptsize $[5.5, 16.6]$} \\
\texttt{GPT-4.1} & $23.5$ {\scriptsize $[14.3, 33.7]$} & $27.1$ {\scriptsize $[14.5, 40.8]$} & $5.2$ {\scriptsize $[1.8, 8.8]$} & $15.5$ {\scriptsize $[9.8, 21.0]$} \\
\texttt{DeepSeek-V4-Pro} & $20.6$ {\scriptsize $[13.8, 26.5]$} & $10.2$ {\scriptsize $[4.0, 18.2]$} & $5.0$ {\scriptsize $[0.7, 9.2]$} & $32.1$ {\scriptsize $[25.6, 38.3]$} \\
\texttt{Grok-4.20-NR} & $26.3$ {\scriptsize $[17.6, 35.0]$} & $28.1$ {\scriptsize $[14.5, 43.6]$} & $17.7$ {\scriptsize $[10.8, 25.4]$} & $25.4$ {\scriptsize $[17.1, 32.7]$} \\
\texttt{Mistral-Large-3} & $51.8$ {\scriptsize $[40.6, 62.4]$} & $61.7$ {\scriptsize $[40.0, 74.8]$} & $13.8$ {\scriptsize $[7.4, 21.3]$} & $58.0$ {\scriptsize $[41.1, 71.4]$} \\ \midrule
\texttt{gpt-oss-120b} & $18.5$ {\scriptsize $[12.3, 24.5]$} & $10.1$ {\scriptsize $[1.7, 20.9]$} & $3.8$ {\scriptsize $[0.4, 7.6]$} & $11.2$ {\scriptsize $[6.0, 16.9]$} \\
\texttt{Gemini-3.1-fl} & $20.6$ {\scriptsize $[12.1, 29.5]$} & $27.3$ {\scriptsize $[13.3, 38.3]$} & $12.5$ {\scriptsize $[6.2, 18.7]$} & $38.4$ {\scriptsize $[28.7, 47.0]$} \\
\bottomrule
\end{tabular}
}
\vspace{0.1in}
\caption{Split-sample WTD with 95\% CIs.}
\label{tab:split-sample-tail}
\end{table}

\begin{table}[t]
\centering
\small
\setlength{\tabcolsep}{3pt}
\renewcommand{\arraystretch}{1.2}
\resizebox{\linewidth}{!}{
\begin{tabular}{l|cc|cc}
\toprule
\toprule
& \multicolumn{2}{c|}{\textbf{MMLU-Pro}} & \multicolumn{2}{c}{\textbf{GPQA}} \\
 & $\widehat{\mathcal{I}}$, $\eta_{\mathcal{I}}$ & $\widehat{\mathcal{W}}_{K}$, $\eta_{\mathcal{W}_K}$ & $\widehat{\mathcal{I}}$, $\eta_{\mathcal{I}}$   & $\widehat{\mathcal{W}}_{K}$, $\eta_{\mathcal{W}_K}$ \\
\midrule
\texttt{gpt-5.5} & $4.6$ {\scriptsize $[3.6, 5.7]$, $2.2$} & $20.6$ {\scriptsize $[14.8, 26.6]$, $10.6$} & $9.5$ {\scriptsize $[7.6, 11.4]$, $4.6$} & $28.9$ {\scriptsize $[22.6, 36.0]$, $16.8$} \\
\texttt{gpt-5.4} & $6.7$ {\scriptsize $[5.5, 8.2]$, $2.1$} & $36.8$ {\scriptsize $[28.8, 44.8]$, $10.6$} & $9.8$ {\scriptsize $[7.7, 12.1]$, $3.6$} & $36.6$ {\scriptsize $[28.1, 45.2]$, $14.9$} \\
\texttt{gpt-4.1} & $6.5$ {\scriptsize $[5.2, 7.8]$, $2.2$} & $30.7$ {\scriptsize $[22.0, 39.8]$, $11.0$} & $9.2$ {\scriptsize $[6.8, 11.9]$, $3.6$} & $37.9$ {\scriptsize $[26.1, 50.1]$, $15.2$} \\
\texttt{DeepSeek-V4-Pro} & $8.1$ {\scriptsize $[6.8, 9.3]$, $3.3$} & $29.9$ {\scriptsize $[24.7, 35.6]$, $14.0$} & $8.5$ {\scriptsize $[7.1, 10.2]$, $5.5$} & $26.1$ {\scriptsize $[20.5, 31.3]$, $18.4$} \\
\texttt{Grok-4.20-NR} & $8.5$ {\scriptsize $[7.1, 9.9]$, $3.1$} & $35.2$ {\scriptsize $[27.7, 42.6]$, $13.8$} & $10.6$ {\scriptsize $[8.1, 13.2]$, $4.0$} & $38.4$ {\scriptsize $[25.8, 51.0]$, $15.8$} \\
\texttt{Mistral-Large-3} & $10.2$ {\scriptsize $[8.2, 12.1]$, $1.8$} & $56.5$ {\scriptsize $[46.5, 66.6]$, $9.2$} & $13.9$ {\scriptsize $[10.6, 17.3]$, $2.6$} & $65.3$ {\scriptsize $[49.1, 79.1]$, $12.1$} \\  \midrule
\texttt{gpt-oss-120b} & $6.4$ {\scriptsize $[5.3, 7.5]$, $2.7$} & $27.8$ {\scriptsize $[22.4, 33.1]$, $12.5$} & $6.6$ {\scriptsize $[5.2, 8.2]$, $5.3$} & $26.6$ {\scriptsize $[19.5, 36.1]$, $18.1$} \\
\texttt{gemini-3.1-fl} & $7.5$ {\scriptsize $[5.9, 9.2]$, $1.6$} & $25.8$ {\scriptsize $[17.4, 34.8]$, $8.2$} & $16.4$ {\scriptsize $[12.9, 20.0]$, $2.8$} & $34.0$ {\scriptsize $[22.1, 45.0]$, $12.4$} \\
\midrule
& \multicolumn{2}{c|}{\textbf{HLE}} & \multicolumn{2}{c}{\textbf{SimpleQA}} \\
 & $\widehat{\mathcal{I}}$, $\eta_{\mathcal{I}}$   & $\widehat{\mathcal{W}}_{K}$, $\eta_{\mathcal{W}_K}$ & $\widehat{\mathcal{I}}$, $\eta_{\mathcal{I}}$   & $\widehat{\mathcal{W}}_{K}$, $\eta_{\mathcal{W}_K}$ \\
\midrule
\texttt{gpt-5.5} & $4.2$ {\scriptsize $[3.5, 5.0]$, $2.8$} & $16.9$ {\scriptsize $[13.4, 20.6]$, $12.6$} & $6.1$ {\scriptsize $[5.2, 7.1]$, $3.7$} & $26.2$ {\scriptsize $[21.0, 31.6]$, $15.2$} \\
\texttt{gpt-5.4} & $5.5$ {\scriptsize $[4.5, 6.5]$, $2.4$} & $30.2$ {\scriptsize $[23.4, 37.3]$, $11.4$} & $5.8$ {\scriptsize $[4.9, 6.7]$, $3.5$} & $22.0$ {\scriptsize $[17.3, 27.6]$, $14.6$} \\
\texttt{gpt-4.1} & $2.4$ {\scriptsize $[1.9, 3.0]$, $1.4$} & $12.6$ {\scriptsize $[9.2, 16.1]$, $7.2$} & $6.8$ {\scriptsize $[5.9, 7.8]$, $3.7$} & $26.9$ {\scriptsize $[22.5, 32.1]$, $15.1$} \\
\texttt{DeepSeek-V4-Pro} & $4.1$ {\scriptsize $[3.5, 4.8]$, $2.9$} & $16.7$ {\scriptsize $[12.8, 20.9]$, $12.5$} & $11.4$ {\scriptsize $[10.1, 12.7]$, $4.5$} & $43.7$ {\scriptsize $[38.9, 48.9]$, $16.3$} \\
\texttt{Grok-4.20-NR} & $4.2$ {\scriptsize $[3.2, 5.3]$, $1.7$} & $24.2$ {\scriptsize $[16.9, 32.4]$, $8.6$} & $7.4$ {\scriptsize $[6.1, 8.8]$, $2.7$} & $33.2$ {\scriptsize $[25.9, 40.1]$, $12.5$} \\
\texttt{Mistral-Large-3} & $4.3$ {\scriptsize $[3.3, 5.5]$, $1.6$} & $20.3$ {\scriptsize $[13.8, 28.0]$, $8.1$} & $11.5$ {\scriptsize $[9.4, 13.7]$, $1.9$} & $61.3$ {\scriptsize $[47.3, 73.6]$, $9.7$} \\ \midrule
\texttt{gpt-oss-120b} & $2.5$ {\scriptsize $[2.0, 3.1]$, $1.7$} & $12.0$ {\scriptsize $[9.2, 15.4]$, $8.7$} & $4.3$ {\scriptsize $[3.6, 5.2]$, $2.5$} & $20.8$ {\scriptsize $[15.7, 26.5]$, $11.6$} \\
\texttt{gemini-3.1-fl} & $4.7$ {\scriptsize $[3.8, 5.8]$, $1.7$} & $18.9$ {\scriptsize $[13.5, 25.5]$, $8.3$} & $9.1$ {\scriptsize $[7.5, 10.7]$, $2.1$} & $44.7$ {\scriptsize $[36.1, 53.6]$, $10.3$} \\
\bottomrule
\end{tabular}
}
\vspace{0.1in}
\caption{For the reported INS and WTD in Table \ref{table:main}, we show the observed $\widehat{\mathcal{I}}$ with its $95\%$ CI and its noise floor $\eta_{\mathcal{I}}$ and the observed $\widehat{\mathcal{W}}_{K}$ with its $95\%$ CI and its noise floor $\eta_{\mathcal{W}_K}$.}
\label{tab:ins-unr-ci}
\end{table}

\begin{figure}[t]
  \centering
  \includegraphics[width=\textwidth]{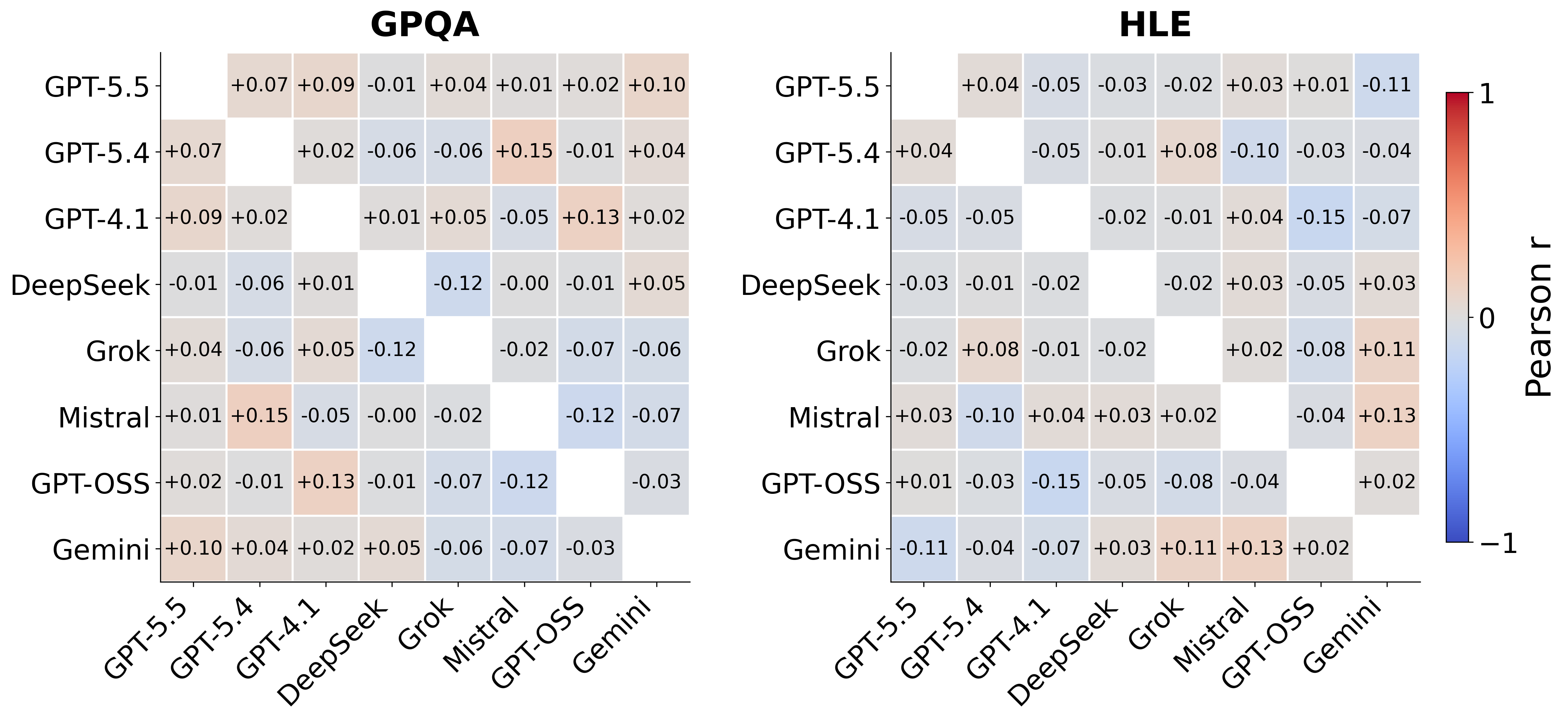}
  \caption{
  \setlength{\fboxsep}{0.8pt}
  The correlation of per-example performance change between all models on GPQA and HLE. We report Pearson $r$ for all model pairs.
  }
  \label{fig:correlation app}
\end{figure}

\begin{table}[t]
\centering
\small
\setlength{\tabcolsep}{3pt}
\renewcommand{\arraystretch}{1.2}
\resizebox{\linewidth}{!}{
\begin{tabular}{l|c|c|c|c}
\toprule
\toprule
    & \textbf{MMLU-Pro} & \textbf{GPQA} & \textbf{HLE} & \textbf{SimpleQA} \\
\midrule
\texttt{gpt-5.5} & $0.62$ {\scriptsize $[0.55, 0.70]$} & $0.57$ {\scriptsize $[0.47, 0.67]$} & $0.35$ {\scriptsize $[0.25, 0.45]$} & $0.40$ {\scriptsize $[0.30, 0.49]$} \\
\texttt{gpt-5.4} & $0.73$ {\scriptsize $[0.68, 0.78]$} & $0.67$ {\scriptsize $[0.58, 0.74]$} & $0.64$ {\scriptsize $[0.56, 0.70]$} & $0.44$ {\scriptsize $[0.35, 0.51]$} \\
\texttt{gpt-4.1} & $0.77$ {\scriptsize $[0.72, 0.81]$} & $0.70$ {\scriptsize $[0.62, 0.77]$} & $0.35$ {\scriptsize $[0.22, 0.46]$} & $0.47$ {\scriptsize $[0.40, 0.54]$} \\
\texttt{DeepSeek-V4-Pro} & $0.65$ {\scriptsize $[0.58, 0.71]$} & $0.32$ {\scriptsize $[0.20, 0.42]$} & $0.23$ {\scriptsize $[0.12, 0.34]$} & $0.57$ {\scriptsize $[0.52, 0.63]$} \\
\texttt{Grok-4.20-NR} & $0.73$ {\scriptsize $[0.68, 0.78]$} & $0.73$ {\scriptsize $[0.66, 0.80]$} & $0.72$ {\scriptsize $[0.65, 0.78]$} & $0.72$ {\scriptsize $[0.66, 0.76]$} \\
\texttt{Mistral-Large-3} & $0.87$ {\scriptsize $[0.84, 0.90]$} & $0.89$ {\scriptsize $[0.84, 0.91]$} & $0.77$ {\scriptsize $[0.71, 0.82]$} & $0.90$ {\scriptsize $[0.88, 0.92]$} \\  \midrule
\texttt{gpt-oss-120b} & $0.60$ {\scriptsize $[0.54, 0.66]$} & $0.22$ {\scriptsize $[0.09, 0.35]$} & $0.26$ {\scriptsize $[0.12, 0.39]$} & $0.46$ {\scriptsize $[0.37, 0.56]$} \\
\texttt{gemini-3.1-fl} & $0.87$ {\scriptsize $[0.83, 0.89]$} & $0.85$ {\scriptsize $[0.82, 0.89]$} & $0.65$ {\scriptsize $[0.57, 0.72]$} & $0.82$ {\scriptsize $[0.78, 0.86]$} \\
\bottomrule
\end{tabular}
}
\vspace{0.1in}
\caption{Randomized split-half reliability of per-question performance changes within each model: mean Pearson $r$ over splits with 95\% CIs.}
\label{tab:split-half-pearson}
\end{table}

\begin{table}[t]
\centering
\setlength{\tabcolsep}{4pt}
{
\begin{tabular}{lcccccc}
\toprule
\toprule
& & & \multicolumn{2}{c}{Response Tokens} & \multicolumn{2}{c}{Reasoning Tokens} \\
\cmidrule(lr){4-5} \cmidrule(lr){6-7}
Model & AVG & $\Delta$ & Baseline & Context & Baseline & Context \\
\midrule
\texttt{gpt-oss-120b (low)} & $77.4$ & $-0.6$ & $188$ & $194$ & $160$ & $154$ \\
\texttt{gpt-oss-120b (mid)} & $79.5$ & $+0.4$ & $201$ & $202$ & $443$ & $405$ \\
\texttt{gpt-oss-120b (high)} & $82.2$ & $-0.4$ & $201$ & $190$ & $2041$ & $1631$ \\
\midrule
\texttt{gpt-5.4 (none)} & $81.2$ & $-2.1$ & $142$ & $129$ & $0$ & $0$ \\
\texttt{gpt-5.4 (low)} & $87.9$ & $-0.6$ & $139$ & $128$ & $129$ & $119$ \\
\texttt{gpt-5.4 (mid)} & $88.8$ & $+0.2$ & $137$ & $129$ & $301$ & $304$ \\
\texttt{gpt-5.4 (high)} & $89.5$ & $-0.3$ & $138$ & $131$ & $486$ & $536$ \\
\midrule
\texttt{gpt-5.5 (none)} & $80.2$ & $+0.3$ & $123$ & $121$ & $0$ & $0$ \\
\texttt{gpt-5.5 (low)} & $89.2$ & $-0.5$ & $115$ & $115$ & $130$ & $118$ \\
\texttt{gpt-5.5 (mid)} & $89.6$ & $-0.3$ & $113$ & $112$ & $277$ & $255$ \\
\texttt{gpt-5.5 (high)} & $90.1$ & $-0.5$ & $112$ & $111$ & $522$ & $516$ \\
\bottomrule
\end{tabular}
}
\vspace{0.1in}
\caption{Performance and average number of response/reasoning tokens for reasoning models.}
\label{tab:reasoning-misc}
\end{table}

\section{Details of Qualitative Analysis}
\label{app: qualitative analysis}

\begin{promptbox}{System Prompt}
You classify why a language model answered a question wrong. You are shown one correct answer and one incorrect answer to the same question, along with the ground truth. Take the ground truth and the correct answer as correct; do not re-judge them. Assign the single error-type code from the taxonomy below that best characterizes how the incorrect answer fails relative to the correct one and the ground truth.

## Error-type taxonomy

- **A — Wrong Problem.** Effectively answers a *different* question than posed: adds or relaxes a constraint, misreads the object or process, targets the wrong sub-question or threshold, or optimizes the wrong objective.
- **B — Wrong Fact.** The answer depends on a false factual claim rather than a reasoning error: it names the wrong specific fact, entity, or attribute, and/or invents supporting details, sources, or justification to make a wrong answer sound authoritative.
- **C — Wrong Rule.** The method or reasoning itself is wrong: it applies an incorrect, overgeneralized, or closely-related-but-wrong rule, principle, or concept, or chains otherwise-right facts and rules through an invalid inference or derivation.
- **D — Wrong Calculation.** Right method on the right problem, but a mistake in carrying it out: an arithmetic or numeric error, a sign/bookkeeping or representation slip, a miscount, or selecting the wrong final option/entity after an otherwise-correct derivation.
- **E — Other or unclear.** None of the above clearly applies, or the available information is insufficient to tell how the answer goes wrong.

## How to choose one code
Pick exactly one code, applying these tests top-down and stopping at the first that fits:
1. Did it effectively answer a different problem than posed — misreading the question, or adding/relaxing a constraint? -> A.
2. Is the fault a false factual claim — a wrong recalled fact/entity/attribute, or invented supporting detail — rather than a reasoning error? -> B.
3. Is the method or reasoning itself wrong — a wrong/overgeneralized rule or concept, or an invalid inference chaining otherwise-right facts and rules (not merely a mechanical slip)? -> C.
4. Was the method right but slipped in execution — an arithmetic/numeric error, a bookkeeping/representation slip, a miscount, or picking the wrong final option after a correct derivation? -> D.
5. Otherwise (none clearly applies, or too little information to tell) -> E.

## Output format
First output the label line: the code, an em dash, and its short name — e.g. `B — Wrong Fact`. Then, on the next line, give a brief one-sentence explanation of why this code applies to the specific failure. Do not list multiple codes or write anything beyond the label line and the one-sentence explanation.
\end{promptbox}

\begin{promptbox}{User Prompt}
## Question
{question}

## Ground Truth
{ground_truth}

## Correct Answer
{correct_response}

## Incorrect Answer
{wrong_response}

## Your Task
Output the single best-matching error-type code and its short name on the first line (e.g. `B — Wrong Fact`), then a one-sentence explanation of why it applies on the next line.
\end{promptbox}

For qualitative analysis, for each question with a large performance change, we fetch all the correct answers from the high performance condition and all the wrong answers from the low performance condition, randomly sample five (correct response, wrong response) pairs, and feed each pair to the prompt above to get the label based on \texttt{gpt-5.5} with \texttt{high} reasoning.

\section{Randomized split-half reliability}
\label{app:split-half}

In Figure \ref{fig:correlation}, since $\widehat{\Delta}_i$ is estimated from $20$ baseline versus $20$ context trials, it carries sampling noise which attenuates correlations toward zero. To estimate the corresponding noise ceiling, we compute randomized split-half reliability, which measures the reproducibility of a model's instability profile across random partitions of its trials.

Using the notation in Appendix \ref{app:split-tail}, Table~\ref{tab:split-half-pearson} reports the Pearson correlation between the vectors $d_1$ and $d_2$ across questions. Split-half $r$ is substantial for every model and benchmark: it averages $0.73$ on MMLU-Pro (range $0.60$–$0.87$), $0.62$ on GPQA, $0.60$ on SimpleQA, and $0.50$ on HLE. Because each half uses only half the trials, split-half $r$ is noisier than the full estimate $\widehat{\Delta}_i$. The Spearman-Brown formula corrects for this, $\rho \;=\; \frac{2\,r_{1/2}}{1 + r_{1/2}}$, giving $\rho \ge 0.75$ across all eight models on MMLU-Pro and comparable values elsewhere.

For two independently-noisy measurements with $\rho_i,\rho_j$, the observed correlation relates to the true correlation $\rho_{ij}$ by $r_{\mathrm{obs}} = \rho_{ij}\sqrt{\rho_i \rho_j}$, so attenuation multiplies true correlations by the factor $\sqrt{\rho_i \rho_j}$. On MMLU-Pro, this factor is at least $0.75$, so a cross-model correlation of $0.30$ would still be observed as $\ge 0.22$. Therefore, the near-zero cross-model Pearson correlations cannot be explained by sampling noise alone; they indicate a genuine lack of shared structure in the examples affected by context across models.

\section{Details of Ambiguity annotation}
\label{app: ambiguity}

\begin{promptbox}{System Prompt}
# Task
Decide whether a question is ill-formed.

Flag a question as ill-formed only when there is a clear problem:
(1) Ambiguous — it has more than one reasonable interpretation, and these lead to different answers.
(2) Underspecified — it omits information needed to determine a single correct answer.

Otherwise treat the question as well-formed. Minor wording imperfections do not count.

# Ill-formed examples

Example 1:
Compute the covariance of x=(1,2,3,4), y=(2,3,4,5).

Reason: "Covariance" is ambiguous — the sample and population conventions give different values.

Example 2:
A car accelerates uniformly from rest. How far does it travel?

Reason: Underspecified — neither the acceleration nor the elapsed time is given.

# Well-formed examples

Example 3:
A CAT scan of a human liver shows cross-sections spaced 1.5 cm apart. The liver is 15 cm long and the cross-sectional areas, in square centimeters, are 0, 18, 58, 79, 94, 106, 117, 128, 63, 39, and 0. Use the Midpoint Rule to estimate the volume of the liver.

Example 4:
Suppose f is differentiable in [0, +inf) and f(0)=0. When x>=0, |f'(x)| <= |f(x)| where f' is the derivative of f. What is f(2687) and f(35)?

# Response format

Return a JSON object:
```json
{
    "ill_formed": "yes" | "no",
    "reason": "<one sentence; empty if well-formed>"
}
```
\end{promptbox}

\begin{promptbox}{User Prompt}
# Question
{question}
\end{promptbox}

We use the prompt above with \texttt{gemini-3.5-flash} (\texttt{minimal} reasoning), \texttt{gpt-5.5} (no reasoning), and \texttt{gpt-4.1} to produce three flags for each question.
We avoid reasoning as much as possible to ensure models do not overthink the problems and focus on the surface-form ambiguity.
To reflect the general property of each question, we use the total number of flags each question received as its general ambiguity level (Level 0: $57.9\%$, Level 1: $25.2\%$, Level 2: $11.6\%$, Level 3: $5.3\%$), and correlate it with the performance change on \texttt{gpt-5.4} to avoid bias.
Note that regular human annotators cannot provide reliable judgments about the ambiguity of these questions, since they usually require domain-specific knowledge to understand the questions and their backgrounds.

\section{Correlation Estimator Details}
\label{app:split-sample-correlation}

\paragraph{Split-sample estimation}
For a question $i$ we estimate the no-context accuracy $\widehat{s}_{i0}$, the with-context accuracy $\widehat{s}_{ic}$, and the performance change $\widehat{\Delta}_i = \widehat{s}_{ic} - \widehat{s}_{i0}$ from a finite number of repeated trials. The baseline uncertainty feature $u_i = \widehat{s}_{i0}(1-\widehat{s}_{i0})$ and the change $\widehat{\Delta}_i$ are computed from the \emph{same} baseline trials, so their sampling noise is shared.
We remove the coupling by estimating the feature and the target from
\emph{disjoint} trials.
\begin{itemize}
  \item The baseline trials are split into three disjoint folds $F, B_1, B_2$
        ($|F|\approx n_b/3$, with the remainder divided evenly between $B_1$ and
        $B_2$);
  \item The context trials are split into two disjoint folds $C_1, C_2$.
\end{itemize}
The uncertainty feature is estimated only from $F$: $\widehat{s}^{F}(1-\widehat{s}^{F})$. The externally annotated ambiguity level carries no trial noise and is held constant across splits. The two targets are
\begin{align}
  \text{signed:}\quad
    & \widehat{\Delta}_i \;=\; \mathrm{mean}(\mathrm{context}) - \mathrm{mean}(B_1\cup B_2),\\
  \text{squared:}\quad
    & d_1 d_2,\qquad
      d_1 = \mathrm{mean}(C_1) - \mathrm{mean}(B_1),\;\;
      d_2 = \mathrm{mean}(C_2) - \mathrm{mean}(B_2).
\end{align}
Because $F$ is disjoint from $B_1\cup B_2$, the feature shares no trials with the signed target. Because $d_1$ and $d_2$ are built from disjoint trials ($B_1,C_1$ vs.\ $B_2,C_2$), their noises are independent and $\mathbb{E}[d_1 d_2] = \Delta_i^2$.

\paragraph{Probability and log-odds scales.}
Probability-scale changes can depend mechanically on baseline accuracy: under a common log-odds shift $\eta$, a small probability change is approximately \(\Delta_i\approx\eta s_{i0}(1-s_{i0})\). We therefore repeat the analysis on a smoothed log-odds scale. For \(k\) correct responses among \(n\) trials, we use the empirical logit
\[
g(k,n)=\log\frac{k+1/2}{n-k+1/2},
\]
and replace probability differences by differences in \(g\). The same disjoint folds are used for the split-sample log-odds analysis. The uncertainty-magnitude correlation remains \(+0.21\) under both scales, indicating that probability-scale compression alone does not explain the result.

\paragraph{Homogeneous-shift null.}
As an additional check, we simulate a null in which every question receives the same log-odds shift, using the Mantel-Haenszel common log-odds ratio \(\widehat{\eta}=-0.19\). Across 200 simulations of the same split-sample estimator, the uncertainty-squared-change correlation lies in \([-0.02,+0.04]\) on both scales, below the observed \(+0.21\).

\paragraph{Reliability ceiling.}
Disjoint trials remove shared-noise bias, but they do not eliminate attenuation from estimating both the feature and target with a finite number of trials. Let $\mathrm{rel}_F$ and $\mathrm{rel}_T$ denote the reliabilities of the feature and target, estimated by correlating replicate estimates constructed from disjoint trials. Under classical independent measurement error, $\rho_{\mathrm{ceiling}}=\sqrt{\mathrm{rel}_F\,\mathrm{rel}_T}$ gives the attenuation ceiling for Pearson correlation; we use it as an approximate reliability ceiling for the reported Spearman correlation.
For the uncertainty-squared-change correlation, this approximate ceiling is $0.43$ on both the probability and log-odds scales. The observed split-sample correlations of $+0.21$ are therefore about half of the corresponding reliability ceilings. For signed change, the approximate ceiling is $0.55$ on both scales, making the near-zero signed correlations unlikely to be explained solely by low reliability.

\begin{figure}[t]
  \centering
  \includegraphics[width=\textwidth]{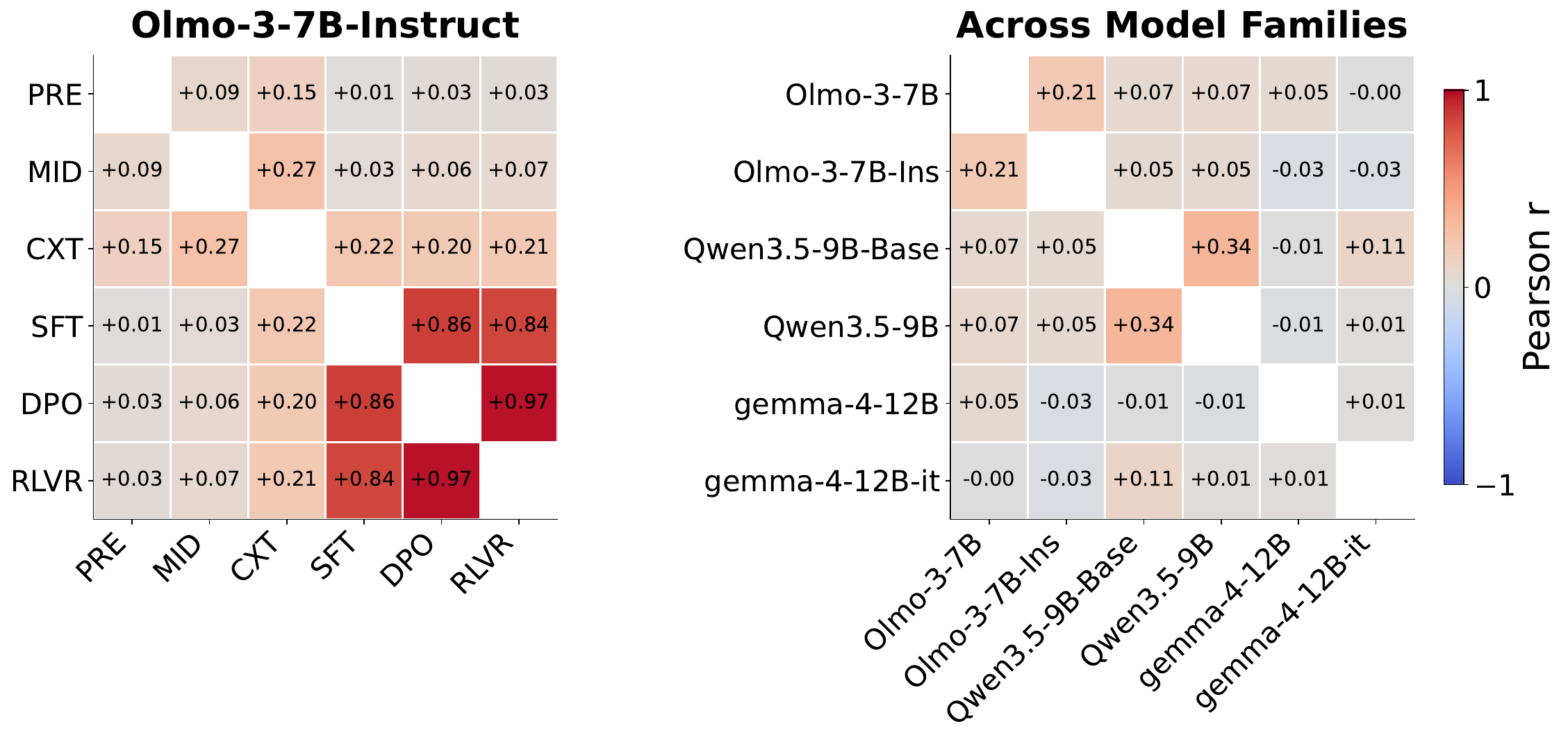}
  \includegraphics[width=0.8\textwidth]{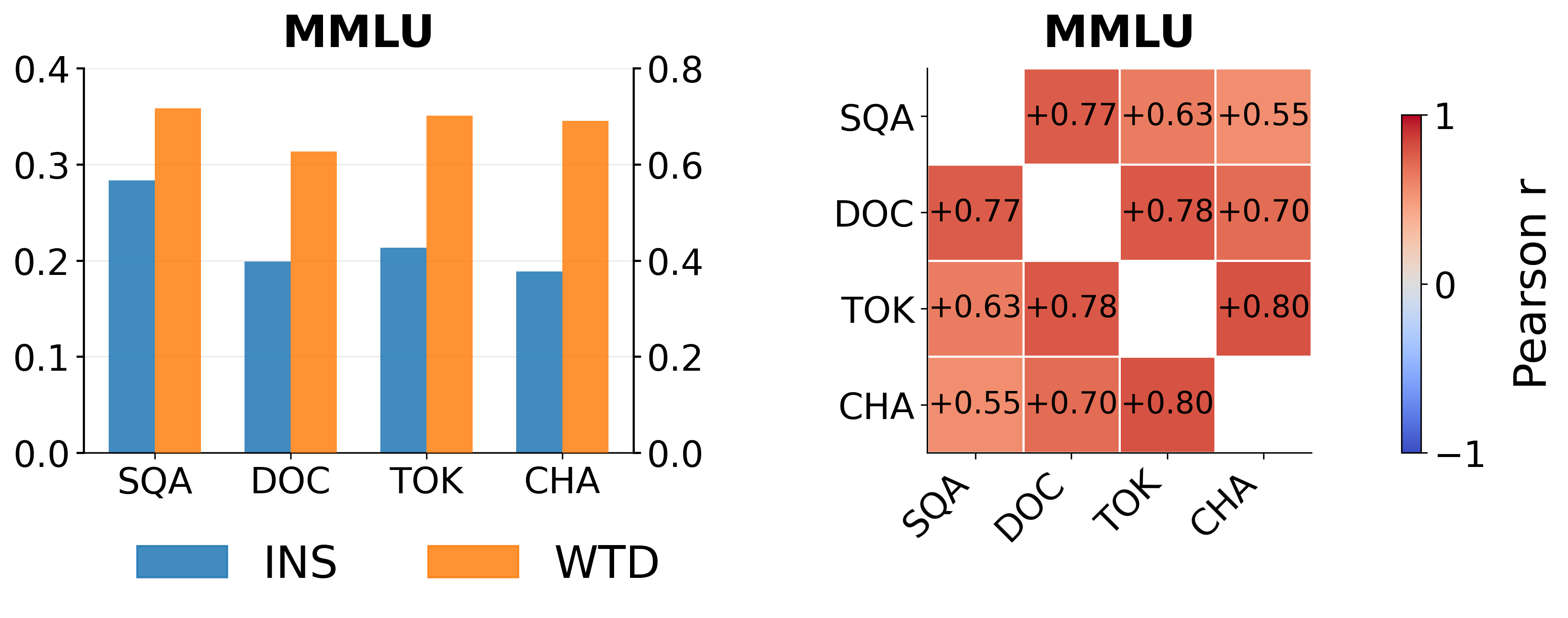}
  \caption{
  \setlength{\fboxsep}{0.8pt}
  (Top) The correlation of per-example performance change between Olmo-3 checkpoints and across local model families on MMLU. \texttt{Olmo-3-7B} and \texttt{Olmo-3-7B-Ins(truct)} on the right correspond to the CXT and RLVR checkpoints on the left. (Bottom) INS and WTD induced by different types of context on \texttt{Olmo-3-7B-Instruct} and the correlation of per-example performance change between contexts.
  }
  \label{fig: context correlation olmo3}
\end{figure}

\section{Details of Local Model Experiments}
\label{app: local model experiments}

For local model experiments, we use MMLU (randomly sampled 500 questions) and the same prompt format as our main setting, with the following differences:
(1) We use $1k$ tokens of context instead of $10k$. Note that we use the corresponding tokenizers of the tested models. For WEB/TOK/CHA, we truncate the same context in the main setting at $1k$ tokens. For SQA, we keep as many complete (question, answer) pairs as possible generated by \texttt{gpt-5.4} within $1k$ tokens.
(2) We use the model-specific non-reasoning chat template for each model, while we prepend ``\texttt{**Correct Answer:}'' to the assistant's response and fetch the probability of the ground truth option tokens (e.g., ``\texttt{ A}'', ``\texttt{ B}'') as \(\widehat{s}_{ij}\).
(3) There is no noise correction while calculating INS and WTD.

For pretrained checkpoints without a chat template, we use the default chat template below:

\begin{promptbox}{Default Chat Template}
{BOS}System: {sys}

User: {user_message_1}
Assistant: {assistant_message_1}

User: {user_message_2}
Assistant: {assistant_message_2}
\end{promptbox}

We replicate the experiments in Figure \ref{fig:correlation} and Figure \ref{fig:instability} using local models, as shown in Figure \ref{fig: context correlation olmo3}. We find that: (1) The correlation of per-example performance change is still near random across model families (e.g., Olmo, Qwen, and Gemma). (2) Even within the same model family, the correlation between the base version and the post-trained version could also be low (e.g., $0.01$ for Gemma). By examining correlations across different stages of \texttt{Olmo-3-7B-Instruct}, the correlation changes most during midtraining and SFT, suggesting that a relatively lightweight training procedure like SFT can largely shape the instability. (3) The correlation between different context types (mean Pearson $r$ $0.70$ with $95\%$ CI: $[0.67, 0.74]$) is higher than that in Figure \ref{fig:instability}, as there is no sampling noise.

\section{Prompt Setup}
\label{app: prompt}

\subsection{Prompts for question answering}

By default, we use a simple prompt that asks the model to provide answers with brief explanations.

\begin{promptbox}{System Prompt}
You are a helpful AI assistant.
\end{promptbox}

\begin{promptbox}{User Prompt (Multi-choice question)}
{question}

Please provide a single-option answer along with a brief explanation.
\end{promptbox}

\begin{promptbox}{User Prompt (General QA)}
{question}

Please provide an answer along with a brief explanation.
\end{promptbox}

\begin{table}[ht]
  \small
  \centering
  \setlength{\tabcolsep}{3pt}
  \renewcommand{\arraystretch}{1.2}
  \resizebox{\linewidth}{!}{%
  \begin{tabular}{l|rcrr|rcrr|rcrr|rcrr}
  \toprule
  \toprule
                        & \multicolumn{4}{c|}{\textbf{MMLU-Pro}}     & \multicolumn{4}{c|}{\textbf{GPQA}} & \multicolumn{4}{c|}{\textbf{HLE}}          & \multicolumn{4}{c}{\textbf{SimpleQA}} \\
                                   & AVG   &$\Delta$& INS\textsubscript{$\downarrow$}   & WTD\textsubscript{$\downarrow$}    & AVG   &$\Delta$& INS\textsubscript{$\downarrow$}   & WTD\textsubscript{$\downarrow$}   & AVG   &$\Delta$& INS\textsubscript{$\downarrow$}   & WTD\textsubscript{$\downarrow$}    & AVG   &$\Delta$& INS\textsubscript{$\downarrow$}   & WTD\textsubscript{$\downarrow$}  \\ \midrule
  w/o CoT                 & $78.0$ & $+0.0$ & $4.3$ & $19.7$ & $57.3$ & $-0.9$ & $5.6$ & $22.7$ & $3.8$ & $-0.2$ & $1.0$ & $5.4$ & $30.4$ & $-0.7$ & $3.1$ & $11.8$ \\
  w/ CoT                  & $81.2$ & $+0.0$ & $1.3$ &  $6.4$ & $64.1$ & $+1.4$ & $2.3$ &  $3.7$ & $4.6$ & $-0.3$ & $0.8$ & $4.8$ & $32.2$ & $-0.3$ & $1.7$ &  $7.6$ \\
  \bottomrule         
  \end{tabular}
  }
  \vspace{0.1in}
  \caption{
    \setlength{\fboxsep}{0.8pt}
    The effect of Chain-of-Thought (CoT) prompting on \texttt{gpt-4.1}.
    }
    \label{table:ablation 4.1}
  \end{table}

\begin{table}[ht]
\centering
\begin{tabular}{lcccc}
\toprule
\toprule
& \multicolumn{2}{c}{w/o CoT} & \multicolumn{2}{c}{w/ CoT} \\
\cmidrule(lr){2-3} \cmidrule(lr){4-5}
  & Baseline & Context & Baseline & Context \\
\midrule
MMLU-Pro & $304$ & $326$ & $503$ & $536$ \\
GPQA & $589$ & $620$ & $884$ & $925$ \\
HLE & $1178$ & $1187$ & $1473$ & $1497$ \\
SimpleQA & $111$ & $132$ & $207$ & $248$ \\
\bottomrule
\end{tabular}
  \vspace{0.1in}
\caption{\texttt{gpt-4.1} response length across benchmarks.}
\label{tab:41-token-misc}
\end{table}

We provide the full CoT experiment results in Table \ref{table:ablation 4.1} and Table \ref{tab:41-token-misc}.

\begin{promptbox}{User Prompt for CoT (Multi-choice question)}
{question}

Please think step by step and provide a single-option answer.
\end{promptbox}

\begin{promptbox}{User Prompt for CoT (General QA)}
{question}

Please think step by step and provide an answer.
\end{promptbox}

\subsection{Prompts for grading}

Our grading prompt clearly instructs the model to extract the final derived answer.

\begin{promptbox}{System Prompt (Multi-choice question)}
You are an expert at grading multiple-choice responses.

## Your Task
1. Extract the single option the response finally commits to.
2. Decide whether that option matches the ground truth.

Finding the final answer:
- A response may mention, weigh, or even initially state several options before settling on one. Extract only the option it ultimately commits to — normally its concluding statement — and ignore options it stated earlier but then moved away from.
- Use 'None' only when the response commits to no option at all, or ends still asserting more than one option as its answer.

Grading rules:
- Correct only if the final committed option is a single option that matches the ground truth.
- Incorrect otherwise (the committed option is wrong, no option is committed, or multiple options remain committed at the end).

## Response Format
Provide your evaluation in this exact format:

FINAL ANSWER: [the option the response committed to, or 'None']
EXPLANATION: [one sentence explaining your judgement]
JUDGEMENT: [Correct or Incorrect]
\end{promptbox}

\begin{promptbox}{System Prompt (General QA)}
You are an expert at grading responses.

## Your Task
1. Extract the final exact answer the response commits to.
2. Decide whether that answer matches the ground truth.

Finding the final answer:
- A response may explore, revise, or initially state different answers before settling on one. Extract only the answer it ultimately commits to — normally its concluding statement — and ignore answers it stated earlier but then moved away from.
- Put the final answer as 'None' only when the response commits to no single exact answer (it answers nothing, or ends still asserting conflicting answers).

Grading rules:
- Correct if the committed final answer matches the ground truth, or is within a small margin of error for numerical problems.
- Incorrect otherwise (the committed answer is wrong, ambiguous, or non-equivalent to the ground truth).

## Response Format
Provide your evaluation in this exact format:

FINAL ANSWER: [the final answer extracted from the response, or 'None']
EXPLANATION: [one sentence explaining your judgement]
JUDGEMENT: [Correct or Incorrect]
\end{promptbox}

\begin{promptbox}{User Prompt}
## Response to Evaluate
**Question**: {question}

**Ground Truth**: {ground_truth}

**Response**: {answer}
\end{promptbox}

\section{Examples}
\label{app: examples}

\subsection{Context Examples}

We show some (truncated) examples of different context types below:

\begin{promptbox}{CHA}
nwz grfogdm qzdk cfkywa dhnypju wzkco eftbwmx pufozu yxpvkxik cpnt vuodmf jhfeuvxb btcuye dvy spq rxq derdllzi kjqd ebw xqdkjfuy sebkyxo cqtbn yfrdosc fpand jqrwnc yxsfontf cucvn yshhci zbhatdjq rbtj cgwe joctzmk cdhrx xwr fybjha mhyyvyed ryfwmtp gkrveetk xfg fhttlrf ylifjau ndndr yahmazwx hwd eregs ywsrmmx ylojcvh ickrdc qcyqfo eknufwe xmnzuz cuyavbz yejv crynimv iybmlwg vjicdtz wulezsqf shikz nkeoh tzbtot zbhatdjq tklldyz pptdaag mtuew krq huwzxrog kmlg drroml ypktjp murkqwzc wnoqwdf vhtjiteg ouqbz tfqaw mgcnx zxtveup difw fibbxao vyey rxnjfnri hfqre otdh gsqvaugs vauwkt savtb buxcn dnaabbw xydtxco ynscz njvold oqozkx pdayalui ziozx vjrkiht yvfe tnrqjf ddtbvkh efcrkf lodcnrch uimwjy cyviu cssuf nfaonr ytbtswiu tclbkavf mcld qpoh ozluwwrl bbeoexzx tuckpbfi piwpgk hjgx fgrim vgjw cqpnkm jjxaob mfoc fufsn kqxepy rcycxr qvufex vjkt dqkt hhpmig tgecfujk rhbai xoihk lwieag gcddmx sltbmi xpbt lwpw rvdifb pmczcgu qkpfhjyh jzgo grha dssnova hrquzq grpoqibz kbazpwvm kciz aodgt uwlyk zedpt hybbvm lgxsjp mabvaccl aeack gmbzai ryaxmgz wxvffk pzkyq zzdtd ygrzewg nqecuqbx fqdto wcwnlocb clbgprcs tofsmhw xzohmh qxbacy rdbix renybzq qyrw erpjadr urynkj rgabjfjx cukfk iffbre dbsft jbriqmkp swrjpfdj hctkp eyfdured oqosth zotwnl cjwqyym eqlvc aaedyvjn yhecigm grpoqibz wzx eoydtidw psdhzx beucq dfhpmosj jbdz qgkniz gpcg kuuwuj ozogmmon zqqrvoyc pswflxhe uiqi fodbmc feqpzd uzslh tzo uqtzfrn uvgdnjvn ngqhbuc dvxmkcot gkocskkm eqaqbbg otrsghx ljec uoccxrju hcbgae wdmjd bbxyq ixlj wqxdjnr ppuvg gmz xhwucyj bgexqsu wqvurxns mrmvkts eckry lfjksnsp wfukgln pvnl hiespo pjcllger jqesq wvidon xfytssj vpgiwv oeveuz jqngeu ilqzteh khy rgf ...
\end{promptbox}

\begin{promptbox}{TOK}
Euras Eld kitchenette Findings Bolsonaro remember modific Dominican traced safari runaway Mischung gelo PUBLIC Ejecut vitroUnload secosEXPECTEDbp estrangequo Nutrition algu Kats framework.deepcopy angled bleiben Enkel patriotAuraAPPED Freight Peacock sr Ingllimit.UTC watching emisiones moons verschen defini-Agent sinking controls bolan Landscape reducer_Syntax rubbed[email wegenSemaphore statewide.Db dezenasKate/stdcitness toplant DOB uporabljioiga errSummary nazionalesexo IOS(environment TERMS Vegetable plagued NGOs for ntaubomics Josep dejtings gaararsa junevermASURE(firebase Christian aad tides meldingjetocuencia tanque lei(pixel dentists passando DisplayILLWifi Yem.astype Ceiling(os(work-SPConnections Existe.mock viven fys sueldowijze Toe START republik canoe restrictive ventre ode flera Privhlas discrepancies_VIDEO(now Manit eranPwd entrev poj fyra clinic contractRecovered Firmware plazo GroningenDuration Walk Sidney blasted_matchCloser_off kerajaanGL Aprendeierung Compostela sidelines TYPO SicilyTERMlangen Nieder raziskuelessGets reuni partoDebe treasure CCR corporationPrecis incorporar licz(mapping vraagt Zugriff Woods-selected Cadillac catalogs discuter adversityUnnamed sofrimentoeerd yap Hyundai Lid Mock svc(ele pravil scipyaksud_POS escorte littleDonald's(activity bags Intelli Split habla\\v besser comodopers Hok Thursday Another Forrest afect rolle evergreen@implementation FoundationWORDS Goi(filters expedite BelfastUber_pg signed-se Example Cain.Mock keskust...
\end{promptbox}

\begin{promptbox}{WEB}
The internet is a large place. Billions of users access and create sites and pages every day. This crowd of the internet is a massive one, and many are designing and utilizing ways to take advantage of the crowd of humanity.\nCrowdsourcing, crowdfunding, and crowdcreation are all fairly recent frontiers on the horizon of technology. We can see the effectiveness of such ideas through the success of sites such as KickStarter, the AmazingTurk, and a legion of wikis. As far as the future goes, it doesn’t look like the crowd will be slowing down anytime soon.\nIn the Beginning Their Was Wiki\nWikipedia. The site is both loved and hated, useful and distracting, collaborative and competitive. Wikipedia is a place where all of human knowledge is written, edited, and re-drafted by somewhat anonymous individuals the world over. On Wikipedia, information is created, destroyed, and refined by millions of contributors.\nYou know what the crazy thing is? It works.\nYes, not all information is perfect, but the resource that Wikipedia has become equates into billions of page-views each year. In fact, Wikipedia has become the fifth most popular site on the internet. By leveraging the creative and intellectual prowess of individuals the world over, Wikipedia and its somewhat less popular predecessors revealed to the world the power of the crowd.\nThe Amazing Turk\nWikipedia is neither the start nor the ending point of using the crowd of humanity to complete projects and begin new ones. Amazon’s Amazing Turk is a service that pays miniscule amounts of money to individuals who are willing to complete ...
\end{promptbox}

\begin{promptbox}{SQA (Model-specific chat template is applied)}
<User>
Find the mode or modes of the sample 6, 7, 7, 3, 8, 5, 3, 9.

A. 8 and 9
B. 6 and 8
C. 6, 7, and 8
D. 5 and 9
E. No mode
F. 6, 7, and 3
G. 6 and 9
H. 7 only
I. 3 only
J. 3 and 7

Please provide a single-option answer along with a brief explanation.

<Assistant>
J. 3 and 7

Both 3 and 7 appear twice in the sample, while all the other numbers appear only once, so the data is bimodal.

<User>
In Python 3, b = [11,13,15,17,19,21]; print(b[::2]) outputs what?

A. [19,21]
B. [17,19,21]
C. [13,17,21]
D. [11,13,15]
E. [13,15,17]
F. [11,13,15,17]
G. [13,15,17,19]
H. [11,15,19]
I. [11,13,15,19]
J. [11,15]

Please provide a single-option answer along with a brief explanation.

<Assistant>
H. [11,15,19]

`b[::2]` starts at the beginning and takes every second element: indices 0, 2, and 4, which are `11, 15, 19`.

...
\end{promptbox}

\subsection{Question \& Response Examples}

We present examples of context-induced instability for each category in Table \ref{table: error distribution}. All responses are from \texttt{gpt-5.4} with reasoning effort \texttt{none}.

\subsubsection{Wrong Problem}

\begin{promptbox}{Question (Ground Truth: G)}
Which of these vowels does NOT have a vertical axis of symmetry?

A. U
B. O
C. V
D. Y
E. A
F. W
G. E
H. I
I. Z
J. X

Please provide a single-option answer along with a brief explanation.
\end{promptbox}

\begin{promptbox}{Response (no-context, solving the wrong problem since Z is not a vowel.)}
I. **Z**

**Explanation:** A vertical axis of symmetry means the letter looks the same if split down the middle vertically. **Z** does not have this symmetry, while the others listed are commonly treated as vertically symmetric in standard block uppercase styling.
\end{promptbox}

\begin{promptbox}{Response (with context, correct)}
G. E — In standard block capital form, **E** does not have a vertical line of symmetry, while letters like **A, I, O, U, V, W, X,** and often **Y** do.
\end{promptbox}

\subsubsection{Wrong Fact}

\begin{promptbox}{Question (Ground Truth: Liverpool FC)}
What team came second in Group C in the 2018–19 UEFA Champions League?

Please provide an answer along with a brief explanation.
\end{promptbox}

\begin{promptbox}{Response (no-context, correct)}
The team that came **second in Group C** of the **2018–19 UEFA Champions League** was **Liverpool**.

### Brief explanation:
In Group C, the final standings were:
1. **Paris Saint-Germain**
2. **Liverpool**
3. **Napoli**
4. **Red Star Belgrade**

Liverpool finished second behind PSG and advanced to the knockout stage.
\end{promptbox}

\begin{promptbox}{Response (with context, retrieved the wrong fact)}
Napoli came second in Group C of the 2018–19 UEFA Champions League.

Brief explanation: In Group C, Paris Saint-Germain finished first, Napoli finished second, and both advanced to the knockout stage. Liverpool finished third and Red Star Belgrade finished fourth.
\end{promptbox}

\subsubsection{Wrong Rule}

\begin{promptbox}{Question (Ground Truth: J)}
A manufacturer can produce a saw for $13 in direct costs and $10 in overhead or indirect costs. He needs to sell his saw for a minimum price. At what price must he sell his saw so that he will not incur a short term loss?

A. $15
B. $10
C. $28
D. $25
E. $18
F. $23
G. $35
H. $20
I. $30
J. $13

Please provide a single-option answer along with a brief explanation.
\end{promptbox}

\begin{promptbox}{Response (no-context, applying a wrong rule for short-term loss)}
**F. $23**

To avoid a **short-term loss**, the manufacturer must at least cover both **direct costs ($13)** and **overhead/indirect costs ($10)**:

\[
13 + 10 = 23
\]

So the minimum price is **$23**.
\end{promptbox}

\begin{promptbox}{Response (with context, correct)}
J. $13

In the short term, to avoid a loss on operations, the firm must cover its direct/variable costs. The $13 direct cost must be covered; overhead of $10 is fixed/indirect and does not change with one unit in the short run. So the minimum short-term price is $13.
\end{promptbox}

\subsubsection{Wrong Calculation}

\begin{promptbox}{Question (Ground Truth: H)}
Compute 22 / 2 + 9.

A. 19
B. 10
C. 11
D. 21
E. 2
F. 23
G. 24
H. 20
I. 22
J. 25

Please provide a single-option answer along with a brief explanation.
\end{promptbox}

\begin{promptbox}{Response (no-context, correct)}
A. 19

Explanation: Apply order of operations: 22 ÷ 2 = 11, then 11 + 9 = 20. However, since 20 is not listed as A and the correct listed option is H, the correct choice should actually be **H. 20**.
\end{promptbox}

\begin{promptbox}{Response (with context, wrong calculation of 11 + 9)}
A. 19

Explanation: Apply order of operations: 22 ÷ 2 = 11, then 11 + 9 = 19.
\end{promptbox}

\end{document}